\documentclass[letterpaper, 10 pt, conference]{ieeeconf}
\IEEEoverridecommandlockouts                              

\usepackage{glossaries}
\usepackage{hyperref}
\usepackage{siunitx}
\usepackage{booktabs}
\usepackage{amsfonts}
\usepackage[pdftex]{graphicx}
\usepackage{subcaption}

\usepackage{algorithm}
\usepackage{algpseudocode}
\usepackage{balance}

\usepackage{multirow}
\usepackage{bm}
\usepackage{tikz}
\usepackage{xcolor}

\usepackage{pst-plot}
\usepackage{pstricks}

\usepackage{soul}

\usepackage{amsmath}

\usepackage{listings}

\usepackage{pgfplots}
\usetikzlibrary{arrows,automata}

\tikzstyle{point} = [fill = green!60!black, circle, scale=0.6]
\tikzstyle{control_point} = [fill = green!60!black, circle, scale=0.6]
\tikzstyle{gnode} = [draw=black,fill = white, circle, scale=0.6]
\tikzstyle{gridstyle} = [step=1cm,gray!60,very thin]

\tikzset{
	ctrlpoint/.style={%
		draw=gray,
		circle,
		inner sep=0,
		minimum width=1ex,
	}
}

\usepackage{pgfplots}
\pgfplotsset{compat=1.15}
\usepackage{mathrsfs}
\usetikzlibrary{arrows}
\usetikzlibrary{calc}
\usetikzlibrary{intersections}
\usetikzlibrary{fit}
\usetikzlibrary{backgrounds,scopes}
\usetikzlibrary{positioning, shapes.geometric, arrows, calc, fit, backgrounds, intersections, angles, decorations.pathreplacing}

\newacronym{rl}{RL}{Reinforcement Learning}
\newacronym{drl}{DRL}{Deep Reinforcement Learning}
\newacronym{mpc}{MPC}{model predictive control}
\newacronym{her}{HER}{Hindsight experience replay}
\newacronym{hgg}{HGG}{hindsight goal generation}
\newacronym{mppi}{MPPI}{Model Predictive Path Integral}
\newacronym{pi}{PI}{Path Integrals}
\newacronym{ddpg}{DDPG}{Deep Deterministic Policy Gradient}
\newacronym{gpu}{GPU}{Graphics Processing Unit}
\newacronym{mdp}{MDP}{Markov Decision Process}
\newacronym{dqn}{DQN}{Deep Q-Networks}
\newacronym{uvfa}{UVFA}{Universal Value Function Approximators}
\newacronym{ebp}{EBP}{Energy-Based Hindsight Experience Prioritization}
\newacronym{cpu}{CPU}{Central processing unit}
\newacronym{mpcc}{MPCC}{Nonlinear model-predictive contouring control}
\newacronym{cbf}{CBF}{Control barrier function}
\newacronym{dnn}{DNN}{Deep neural networks}
\newacronym{naf}{NAF}{Normalized Advantage Function}
\newacronym{ctr}{CTR}{Colliding-transition Relabeling}
\newacronym{ddp}{DDP}{Differential Dynamic Programming} 

\renewcommand{\baselinestretch}{0.985}
\begin{document}
%
\title{\LARGE \bf Safety Guaranteed Manipulation Based on Reinforcement Learning Planner and Model Predictive Control Actor}
%
%
%

\author{Zhenshan~Bing$^{1}$,
        Aleksandr~Mavrichev$^{1}$,
        Sicong~Shen$^{1}$,
        Xiangtong~Yao$^{1}$,
        Kejia~Chen$^{1}$,
        \\
        Kai~Huang$^{2}$,
        and~Alois~Knoll$^{1}$
\thanks{$^{1}$Z. Bing, A. Mavrichev, S. Shen, X. Yao, K. Chen, and A. Knoll are with the Department
of Informatics, Technical University of Munich, Germany.
}
\thanks{$^{2}$K. Huang is with the School of Data and Computer Science, Sun Yat-sen University, China.}
}

%
%

\markboth{Journal of \LaTeX\ Class Files,~Vol.~14, No.~8, August~2015}%
{Shell \MakeLowercase{\textit{et al.}}: Bare Demo of IEEEtran.cls for IEEE Journals}
%



\maketitle

\begin{abstract}
\

Deep reinforcement learning (RL) has been endowed with high expectations in tackling challenging manipulation tasks in an autonomous and self-directed fashion. 
Despite the significant strides made in the development of reinforcement learning, the practical deployment of this paradigm is hindered by at least two barriers, namely, the engineering of a reward function and ensuring the safety guaranty of learning-based controllers. 
In this paper, we address these challenging limitations by proposing a framework that merges a reinforcement learning \lstinline[columns=fixed]{planner} that is trained using sparse rewards with a model predictive controller (MPC) \lstinline[columns=fixed]{actor}, thereby offering a safe policy.
On the one hand, the RL \lstinline[columns=fixed]{planner} learns from sparse rewards by selecting intermediate goals that are easy to achieve in the short term and promising to lead to target goals in the long term.
On the other hand, the MPC \lstinline[columns=fixed]{actor} takes the suggested intermediate goals from the RL \lstinline[columns=fixed]{planner} as the input and predicts how the robot's action will enable it to reach that goal while avoiding any obstacles over a short period of time.
We evaluated our method on four challenging manipulation tasks with dynamic obstacles and the results demonstrate that, by leveraging the complementary strengths of these two components, the agent can solve manipulation tasks in complex, dynamic environments safely with a $100\%$ success rate.
Videos are available at \url{https://videoviewsite.wixsite.com/mpc-hgg}.
\end{abstract}


%
\IEEEpeerreviewmaketitle

\section{Introduction}

Deep reinforcement learning (RL) has been widely used to solve complex decision-making tasks in robotics, such as controlling robotic arms to perform manipulation tasks \cite{9466373}, generating agile locomotion gaits for legged robots \cite{hwangbo2019learning}, and planning trajectories for autonomous vehicles \cite{kollar2008trajectory}. 
As there is no modeling computation or optimization involved, RL-based methods are superior to traditional control methods in solving long-horizon planning and dynamic tasks in a timely manner.
However, RL methods are constantly facing two major challenges, namely, requiring handcrafted reward functions that are tailored to individual tasks and lacking rigorous guarantee to ensure the safety of operations.

For the first challenge, in most complex robotic tasks, where a concrete representation of efficient
or even admissible behavior is unknown, it is extremely difficult and time-consuming to design an adequate task tailored reward, thereby making this strategy impractical for wide robotic applications of RL.
One promising concept to address the reward engineering problem is to use a binary reward to simply indicate the completion of the task based on its success or failure condition.
This kind of binary reward is also known as a sparse reward and is easy to derive from task definition with minimum effort.

\begin{figure}[t]
	\input{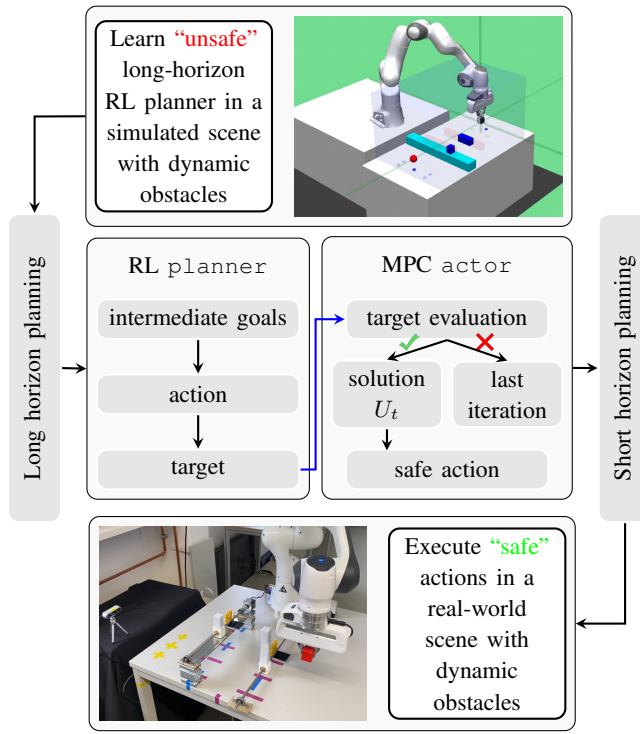}
	\caption{Overview of the proposed MPC-HGG framework. The RL \lstinline[columns=fixed]{planner} performs long horizon planning by proposing intermediate goals. The MPC \lstinline[columns=fixed]{actor} performs short horizon planning to ensure the safety of the action by avoiding dynamic obstacles.}
	\label{fig:framework}
\end{figure}

Although a sparse reward is easy to derive from task definition with minimum effort, RL algorithms that support sparse rewards usually suffer from bad
learning efficiency.
This is because the sparse reward only delivers shallow and
insufficient information during training, which can limit the performance of the RL algorithm.
\acrfull{her}~\cite{her} is one of fundamental methods which improves the success of off-policy RL algorithms in multigoal RL problems with sparse rewards.
The core idea behind HER is to train an agent using handcrafted, easy-to-achieve intermediate goals, and then gradually increase the difficulty of the goals. To achieve this, HER constructs hindsight goals from previously achieved states, replays known trajectories with these hindsight goals, and uses the results to train a goal-dependent value function.
An extremely useful extension to HER is \acrfull{hgg} algorithm, which can generate more meaningful hindsight goals in the direction of the desired goal and accelerate the learning of distant targets~\cite{hggalg}.
However, HGG shows poor performance in the presence of obstacles since distance norm measure for the
hindsight goals distribution does not account for the occupied area.

For the second challenge, learning methods are always criticized by their indistinct interpretability and suffering its adaptability to environments with uncertainty. 
Take manipulation tasks for example, although robotic arms are mostly operated in a relatively enclosed space, the unpredictable movement of human operators or a sudden invasion of unknown objects will pose great safety threat to the completion of the task or even for life and property.
As an effective control approach, the \acrfull{mpc} is extensively used for path planning and collision avoidance~\cite{mpc:overviewspringer, 9812160}.
Model predictive control requires a dynamic model of the environment to predict the future states and optimise control over a finite time horizon.
This means that MPC provides online trajectory optimization with the ability to rapidly react to changes in the environment.

To overcome these challenges, we propose a control framework that combines the advantages of RL-based methods in long-horizon planning and traditional model-based control methods in safety guaranteed performance (see Figure \ref{fig:framework}). 
Specifically, the HGG algorithm is used as the high-level \lstinline[columns=fixed]{planner} to propose intermediate goals for solving a long-horizon task.
While the MPC algorithm is used as an \lstinline[columns=fixed]{actor} to execute the sub-task that takes the intermediate goal proposed by the \lstinline[columns=fixed]{planner} as the input.
We show that the combination of the both methods can guarantee collision avoidance in the randomized dynamic environments.
Our results demonstrate how the MPC \lstinline[columns=fixed]{actor} can effectively address the issue of unreliable safety in a RL policy and how the RL policy can guide MPC through the infinite horizon.

Our contributions to the literature are summarized as follows.
First, we propose an RL \lstinline[columns=fixed]{planner} that utilizes the hindsight goal generation algorithm to generate intermediate goals for long-term manipulation tasks. We enhance the collision avoidance capability of the HGG algorithm by introducing a multi-objective sparse reward concept. 
This reward function incentivizes not only reaching the goal, but also avoiding any obstacle collision, all while minimizing the required engineering effort.
Second, we formulate an MPC \lstinline[columns=fixed]{actor} that can optimize the trajectory of a robotic arm to reach to the goal proposed by the RL \lstinline[columns=fixed]{planner} and avoid colliding with any obstacle over finite horizon.
Last, we show the proposed framework is able to solve complex manipulation tasks effectively and safely in the simulation and the real world.
The controller solves all the tasks with a success rate of $100\%$ and attains a real-time performance with less than $3$ ms per timestep.  

\section{Related Works}

This section aims to provide a brief overview of the literature that explores the combination of model predictive control with reinforcement learning algorithms.

A closely related work by \cite{britogompc} introduces the GO-MPC
algorithm for autonomous navigation in crowded scenarios.
The authors propose a pre-trained on-policy RL algorithm to provide long-term sub-goals to the local motion planner based on MPC.
Hansen et al. \cite{td3mpc} incorporate MPC and twin delayed deep deterministic policy gradient (TD3) to improve sample efficiency via model learning.
The authors jointly train a terminal value function and a latent dynamics model, which are utilized for local trajectory optimization and global guidance, respectively. By leveraging the latent space representation, the learned model can reduce the state space by eliminating irrelevant features, such as background noise and shading.
To address the problem of high computational costs over a long planning horizon, Negenborn et al. \cite{NEGENBORN2005354} propose
to utilize the learned value function in the cost function of a conventional MPC controller.

Bhardwaj et al. \cite{mpc:qlearn} propose a novel approach to capitalize on the advantages of both model-free and model-based methods. Specifically, they propose a Model Predictive Control (MPC) based Q-Learning algorithm that utilizes local optimization to improve the value function and demonstrate faster learning speeds with fewer system interactions. This method provides a promising alternative that combines the strengths of both model-based and model-free methods for control tasks.
%
%
Greatwood et al. \cite{mpc-greatwood} employ MPC to regulate a quadrotor micro air vehicle (MAV) using guidance from a RL policy.
The approach is designed to operate under the assumption of only rectangular obstacles and employs two distinct MPCs for each axis to effectively control the MAV.
Xue et al. \cite{ddpgmpc:xue} leverage MPC within the deep deterministic policy gradient (DDPG) algorithm to predict the trajectory of dynamic obstacles. 
The proposed approach employs a complex reward function that comprises target attraction, obstacle repulsion, collision penalty, and reward for reaching the target. 
This technique demonstrates promise in navigating dynamic environments with obstacles.

Another interesting approach was introduced by \cite{rlmpc:google},
in which the authors suggest learning a control policy directly in the real-world environment. 
The approach utilizes an additional safe policy that can be triggered to return the robot to a safe state during training. An approximated dynamics model is then used to decide when to revert back to the learning policy to continue the training. 
MPC is employed to achieve more stable operation due to the reduced action space and to avoid direct control of motor torques. This methodology offers a promising technique for training control policies in real-world environments while ensuring safety.

\section{Preliminaries}

\subsection{Goal-Conditioned RL} 
In goal-conditioned RL, an agent interacts with its environment to reach some goals, which can be modeled as a goal-conditioned Markov decision process (MDP) with a state space $\mathcal{S}$, an action space $\mathcal{A}$, a goal space $\mathcal{G}$, a probabilistic transition function $P : S \times \mathcal{A} \rightarrow \mathcal{S}$, a reward function $r_g: \mathcal{S} \times \mathcal{A} \rightarrow \mathbb{R}$, and a discount factor $\gamma$.  
The agent's action $a_t$ is defined by a 
probabilistic
policy $\pi(s_t||g)$ at every time step $t$, given by the current state $s_t$ and the goal $g$
(we use $||$ as a symbol for concatenation into $\mathcal{S} \times \mathcal{G}$).
The goal is to find a policy that can maximize the expected curriculum reward starting from the initial state sampled from the initial state distribution $s \in S_0$, which is defined as 
\begin{equation}
	\begin{split}
		V^{\pi}(s||g) =  \mathbb{E}_{s_0, a_t \sim \pi(s_t||g),\, s_{t+1} \sim P(s_t, a_t)}\big[\sum_{t=0}^{\infty} \gamma^t r_g(s_t, a_t)\big]\text{.}
	\end{split}
	\label{eq:V}
\end{equation}

\subsection{Hindsight Experience Replay}

Hindsight Experience Replay (HER \cite{her}) is an RL algorithm specifically engineered for goal-oriented tasks characterized by sparse rewards, which are often difficult for agents to learn efficiently. Despite its simplicity, HER has been demonstrated to be a highly effective method for improving agent performance in these challenging scenarios.
HER is designed to enhance learning efficiency by using a relabeling approach that exploits the idea that experiences that are uninformative for a given goal may still contain valuable information for other goals.
HER assumes that, in a multi-goal RL task with sparse rewards, each goal $g$ is associated with a predicate $f_g: \mathcal{S} \rightarrow \{0,1\}$. 
Once the agent reaches a state $s$ that satisfies $f_g(s) = 1$, it is considered that the goal has been achieved.
The reward function is defined as sparse if it satisfies $r_g(s,a) = -[f_g(s)=0]$. This implies that until the agent reaches the goal, it continuously receives negative rewards.
%
%
In HER, each transition $(s_t || g, a_t, r_t, s_{t+1}|| g)$ is not only stored with the original episode goal $g$, but also with a subset of hindsight goals $g'$ as $(s_t || g', a_t, r_t, s_{t+1}|| g')$. 
As a result, when replaying the resulting transitions $(s_t || g', a_t, r_t, s_{t+1}|| g')$, the agent is more likely to encounter informative rewards. 
An interpretation of HER is that it acts as an implicit curriculum, focusing initially on simpler intermediate goals and subsequently progressing towards more challenging goals that are nearer to the ultimate target goals.

\section{Methodology}

This section first gives an overview of our proposed algorithm MPC-HGG.
Then we explain the RL \lstinline[columns=fixed]{planner} and the MPC \lstinline[columns=fixed]{actor} in detail.
Finally, we summarize MPC-HGG with its pseudocode.

\subsection{Overview}
The overall architecture of the MPC-HGG algorithm is shown in Figure \ref{fig:framework}. 
The algorithm is briefly explained in two phases as follows.
\begin{enumerate}
	\item In the first stage, we design a RL \lstinline[columns=fixed]{planner} that can solve complex, long-horizon planning manipulation tasks via a curriculum learning approach.
	This RL controller can adapt itself to multi-goal tasks, but is not able to guarantee the safety of the proposed action.
	\item In the second stage, we develop a MPC \lstinline[columns=fixed]{actor} that can provide safe actions to reach to an intermediate goal in a short planning horizon. 
	The intermediate goals are suggested by the RL \lstinline[columns=fixed]{planner} from stage one.
\end{enumerate}


\subsection{RL Planner}

\begin{figure}[t!]
	\centering

\begin{tikzpicture}[thick,scale=0.9, every node/.style={scale=0.9}]

\definecolor{color0}{rgb}{0.5,0,1}
\definecolor{color1}{rgb}{0.280392156862745,0.338158274815817,0.985162233467507}
\definecolor{color2}{rgb}{0.0607843137254902,0.636474236147141,0.941089252501372}
\definecolor{color3}{rgb}{0.166666666666667,0.866025403784439,0.866025403784439}
\definecolor{color4}{rgb}{0.386274509803922,0.984086337302604,0.767362681448697}
\definecolor{color5}{rgb}{0.613725490196078,0.984086337302604,0.641213314833578}
\definecolor{color6}{rgb}{0.833333333333333,0.866025403784439,0.5}
\definecolor{color7}{rgb}{1,0.636474236147141,0.338158274815817}
\definecolor{color8}{rgb}{1,0.338158274815818,0.17162567916636}
\definecolor{color9}{rgb}{1,1.22464679914735e-16,6.12323399573677e-17}

\begin{axis}[
tick align=outside,
tick pos=left,
x grid style={white!69.0196078431373!black},
xmin=-0.549542961608775, xmax=4,
xtick style={color=black},
y grid style={white!69.0196078431373!black},
ymin=-10, ymax=10,
ytick style={color=black},
hide axis
]
\addplot [draw=green!50!black, fill=color1, mark=*, only marks]
table{%
x  y
0 0
0.1 1
0.2 2
0.3 3
-0.1 2
-0.2 3
-0.3 1
0.1 -1
-0.2 -2
0.3 -3
};
\addplot [draw=green!50!black, fill=color3, mark=*, only marks]
table{%
x  y
1 0
1.1 2.2
1.2 2.8
1.3 1.2
0.9 1.3
0.8 2.3
0.7 2.7
1.1 -3.2
0.8 -1.8
1.3 -1.2
};
\addplot [draw=green!50!black, fill=color5, mark=*, only marks]
table{%
x  y
2 0
2.1 2.4
2.2 3.2
2.3 1.6
1.9 2.3
1.8 1.3
1.7 2.0
2.1 -3.6
1.8 -2.8
2.3 -2.2
};
\addplot [draw=green!50!black, fill=color6, mark=*, only marks]
table{%
x  y
3 0
3.1 2.0
3.2 -3.2
3.3 -1.6
2.9 -2.3
2.8 1.5
2.7 -2.0
3.1 3.6
2.8 -2.3
3.3 2.2
};
\addplot [draw=green!50!black, fill=color9, mark=*, only marks]
table{%
x  y
4 0
4.1 1.8
4.2 3.2
4.3 1.6
3.9 2.1
3.8 1.7
3.7 -2.2
4.1 2.6
3.8 -2.0
4.3 -2.2
};
\end{axis}

\node (rect1) at (7,3) [draw=red, fill=red!10!white, minimum width=0.2cm,minimum height=0.2cm] {};
\node (rect2) at (0,3) [draw=blue, fill=blue!10!white, minimum width=0.2cm,minimum height=0.2cm] {};

\node (goal) [red] at (6, 1) {final goal};
\draw [-stealth] (goal) -- (rect1);

\node (goal) [black] at (1, 1) {{initial goal}};
\draw [-stealth] (goal) -- (rect2);

\node (int_goal) [blue] at (3.5, 1) {{intermediate goals}};
\draw [-stealth] (int_goal) -- + (0, 2);
\draw [-stealth] (int_goal) -- + (-1.5, 2);
\draw [-stealth] (int_goal) -- + (1.5, 2);

\draw[-stealth] (1.5, 4.2) -- node[above, blue] {guided learning with intermediate goals} (5.5, 4.2);
\end{tikzpicture}
	\caption{Visualization of the intermediate goal distributions generated by the hindsight goal generation.}
	\label{fig:hgg_concept}
\end{figure}
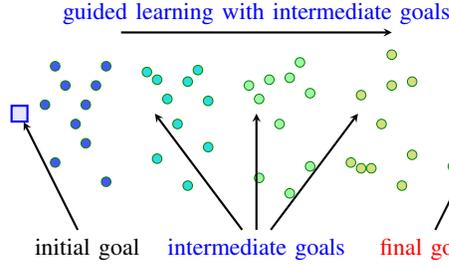

Inspired by HER \cite{her} and HGG \cite{hggalg}, we train the RL \lstinline[columns=fixed]{planner} in the following fashion.
The episode starts with sampling an initial state - goal pair $(s_0, g)$ from an initial state distribution $\mathcal{S}_0$ and a target goal distribution $\mathcal{G}_T$.
A state $s$ can be mapped to a goal $g_s$ by $g_s = m(s)$. 
In the beginning of the learning stage, the exploration is random since no meaningful policy has been developed yet. 
Exploration naturally starts from $s_0 \sim \mathcal{S}_0$, thus goals that are close
to $m(s_0)$ are reached more easily.
The agent can leverage the generalization capabilities inherent in neural networks, enabling it to extrapolate from past experiences and extend its ability to achieve goals similar to those previously reached.
The idea is illustrated in Figure \ref{fig:hgg_concept}.


The distribution $\mathcal{T} ^{*}: \mathcal{G} \times \mathcal{S} \to \mathbb{R}$ determines how they are sampled.
Instead of optimizing the value function $V^{\pi}$ with difficult target goals, which carries the risk of being too far from the known goals, we try to optimize with a set of intermediate goals sampled from $\mathcal{T}$.
On the one hand, the goals contained in $\mathcal{T}$ should be easy to reach, which requires a high $V^{\pi} (\mathcal{T})$.
On the other hand, goals in $\mathcal{T}$ should be close enough to $\mathcal{T^{*}}$ to be challenging for the agent.

Inspired by HGG \cite{hggalg}, a guided schedule for selecting suitable intermediate goals $g' \in \mathcal{G}$ that will be used by the agent instead of $g \in \mathcal{G}_{T}$. 
This will guide the agent to learn from easy to difficult, so it can learn to reach the goals from $\mathcal{G}_{T}$ gradually. 
Therefore, it is necessary to find a substitute distribution $\mathcal{T}: \mathcal{G} \times \mathcal{S} \to \mathbb{R}$ which chooses appropriate intermediate goals. 
On the one hand, such goals must be close to goals that the agent can already reach, and on the other hand, they still have some distances to achieved goals so that the agent learns something new to approach the final goal. 
This trade-off can be formalized as 
\begin{equation}
	\label{eq:optimization}
	\max_{\mathcal{T}, \pi} V^{\pi} (\mathcal{T}) - L \cdot \mathcal{D} (\mathcal{T^{*}}, \mathcal{T}) \text{.}
\end{equation}
The Lipschitz constant $L$ is treated as a hyper-parameter. 
In practice, to select these goals, we first approximate $\mathcal{T}^{*}$ by taking $K$ samples from $\mathcal{T}^{*}$ and storing them in $\hat{\mathcal{T}}^{*}$. 
Then, for an initial state and goal $(\hat{s}_0^i,\hat{g}^i) \in \hat{\mathcal{T}}^{*}$, we select a trajectory $\tau = \{s_t \}_{t=1}^T$ that minimizes the following function:
\begin{equation}
	\label{eq:weights_hgg}
	\begin{aligned}
		w(\hat{s}_0^i,\hat{g}^i,\tau) := {} & c \|m(\hat{s}_0^i) - m(s_0) \| \\ 
		& + \min_{s_t \in \tau} \left(\| \hat{g}^i - m(s_t) \| - \frac{1}{L} V^{\pi} ((s_0 || m(s_t))\right) \text{.}
	\end{aligned}
\end{equation}
$c > 0$ provides a trade-off between 1) the distance between target goals and 2) the
distance between the goal representation of the initial states.
Finally, from each of the $K$ selected trajectories $\tau^i$, the hindsight goal $g^i$ is selected from the state $s_t^i \in \tau^i$, that minimized \eqref{eq:weights_hgg}. More formally,
\begin{equation}
	\label{eq:goal_hgg}
	g^{i} := \mathop{\text{arg min}}_{s_t \in \tau} \left(\| \hat{g}^i - m(s_t)\| - \frac{1}{L} V^{\pi} ((s_0 \| m(s_t))\right) \text{.}
\end{equation} 

Through experimentation, it was observed that the agent was able to achieve the target goal despite encountering obstacles and colliding during task execution, which should be punished as we expect the agent to complete the task while avoiding collisions.
Prior work addresses this issue by creating a dense reward signal linked to specific obstacle measurements to ensure collision-free movement.
However, designing an well-designed reward tailored to a specific task is challenging and demands time and effort.

We suggest a way to balance the advantages of using sparse rewards with the need for task-oriented behavior. 
Our solution is a multi-objective conditioned binary reward function that assigns different magnitudes to each objective.
Specifically, the multi-objective sparse reward is defined as 
\begin{equation}
	r_g(s) := 
	\begin{cases}
		\eta, & \text{if collision }\\
		0, & \text{if} \, f_g(s) = 1\\
		-1, & \text{otherwise.}
	\end{cases}
	\label{eq_sparse_reward}
\end{equation}
The reward $\eta$, as a hyperparameter, is a constant negative value that is customized to prevent the collision.
If the agent encounters an obstacle, it will receive a reward of $\eta < -1$.
However, if the agent does not collide with any obstacles but still fails to reach the goal, it will receive a reward of $-1$.
The agent will only receive a reward of $0$ when it successfully reaches the goal.
The pseudocode of the HGG algorithm is shown in Algorithm \ref{alg:hgg}.

\begin{algorithm}[t!]
	\caption{Hindsight Goal Generation (HGG)}\label{alg:hgg}
	\small{
		\begin{algorithmic}[1]
			
			\State Given: 
			\begin{itemize}
				\item An off-policy RL algorithm $\mathbb{A}$,\Comment{e.g. DDPG}
				\item A strategy $\mathbb{S}$ for sampling goals for replay, \Comment{e.g. $\mathbb{S}(s_0,...,s_T) = m(s_T)$}
				\item A set of reward functions $r_g: \mathcal{S} \times \mathcal{A} \rightarrow \mathbb{R}$ \Comment{e.g. $r_g(s, a) = - \mid f_g(s)==0 \mid $}
			\end{itemize} 
			\State Initialize $\mathbb{A}$
			\State Initialize replay buffer $R$
			
			\For{$iteration$}
			\State Construct a set of $M$ intermediate tasks $\{(\hat{s}_0^i, g^i)\}_{i=1}^M$: \Comment{HGG}
			\begin{itemize}
				\item Sample target tasks $\{(\hat{s}_0^i, \hat{g}^i)\}_{i=1}^K \sim \mathcal{T}^*$ 
				\item Find $K$ distinct trajectories $\{\tau^i\}_{i=1}^K$ that together minimize \eqref{eq:weights_hgg} \Comment{weighted bipartite matching}
				\item Find $M$ intermediate tasks $(\hat{s}_0^i, g^i)$ by selecting and intermediate goal $g^i$ from each $\tau^i$ 
			\end{itemize}
			\For{$episode = 1,M$}
			\State $(s_0,g) \leftarrow (\hat{s}_0^i, g^i)$ \Comment{hindsight goal-oriented exploration}
			\For{$t = 0, T-1$}
			\State Sample an action $a_t$ using the policy from $\mathbb{A}$ with noise:
			\begin{equation}
				a_t \leftarrow \pi(s_t \parallel g) + \mathcal{N}_t 
			\end{equation}
			\State Execute action $a_t$ and observe new state $s_{t+1}$
			\EndFor
			
			\For{$t = 0, T-1$}
			\State $r_t := r_g(s_t,a_t)$
			\State Store transition $(s_t \parallel g,\; a_t,\; r_t,\; s_{t+1} \parallel g)$ \Comment{DDPG experience replay}
			\State Sample a set of additional goals for replay $G:=\mathbb{S}(current\, episode)$
			\For{$g' \in G$}
			\State $r':=r_{g'}(s_t, a_t)$
			\State Store the transition $(s_t \parallel g',\; a_t,\; r',\; s_{t+1} \parallel g')$ in $R$ \Comment{HER}
			\EndFor
			\EndFor
			\EndFor
			\For{$t = 1, N$}
			\State Sample a minibatch $B$ from the replay buffer $R$ \Comment{HER or EBP}
			\State Perform one step of optimization using $\mathbb{A}$ and minibatch $B$ \Comment{DDPG}
			\EndFor
			\EndFor
			
		\end{algorithmic}
	}
\end{algorithm}

\subsection{MPC Actor}

Model predictive control, which is model-based method for trajectory planning, uses a
dynamic model of the system to predict the future position of the agent and optimizes the sequence of the control inputs to achieve lower costs.

\subsubsection{Definition}

Our proposed model is designed based on the idea that modern industrial robots have optimization algorithms that can quickly convert a desired location into specific torque rates for each joint of the robot arm. 
This allows for precise movement of the arm to the desired location.
As the action space used in the RL \lstinline[columns=fixed]{planner} is the gripper's coordinates $[x, y, z]$ of the robotic arm, we use a simple point-mass model for the MPC \lstinline[columns=fixed]{actor}:
\begin{equation}
	\label{eqn:mpcmodel}
	\begin{aligned}
		\dot x = v_x\text{,} \quad \dot v_x = \dfrac{F_x}{m}\\
		\dot y = v_y\text{,} \quad \dot v_y = \dfrac{F_y}{m}\\
		\dot z = v_z\text{,} \quad \dot v_z = \dfrac{F_z}{m}\\
	\end{aligned}
\end{equation}
We define the state vector as $x = [x,y,z, v_x, v_y, v_z] \in \mathcal{X} = \mathbb{R}^6$ and
the control input vector as $u = [F_x, F_y, F_z, \xi] \in \mathcal{U} = \mathbb{R}^4$.
$\xi$ is used to soften the hard constraints.
$m$ is the mass of the manipulable object.

\subsubsection{Formulation}\label{subsection:mpcform}
We define the problem as non-convex, finite-time nonlinear optimal control with horizon length $N$ and takes the following form:
\begin{equation}
	\label{eqn:mpccostform}
	\begin{aligned}
		\text{minimize} \quad & \sum_{k=1}^{N} f_k(z_k, p_k) && \quad \text{cost function}\\
		\text{subject to} \quad & z_1(\mathcal{I}) = z_{\mathrm{init}} && \quad \text{initial equality} \\
		& \underline{z}_k \leq z_k \leq \bar{z}_k && \quad \text{upper-lower bounds} \\
		& \underline{h}_k \leq h_k(z_k, p_k) \leq \bar{h}_k && \quad \text{nonlinear constraints}
	\end{aligned}
\end{equation}
where $z_k \in \mathcal{U} \times \mathcal{X} \in \mathcal{Z}=\mathbb{R}^{10}$ is a stage variable which stacks the input and
differential state variables together. 
$p_k \in \mathcal{G} \times \mathcal{O}_{i,k},\: \forall i \in \{1,\dots,N_o\}$
contains
the real-time data, such as current goal and obstacles.
$f_k(z_k, p_k)$ represents the cost function, which should be minimized during the optimization process, and $h_k(z_k, p_k)$ represents the nonlinear constraint function used for collision avoidance.
For each of these variables, the respective inequalities must be satisfied during each optimization step.

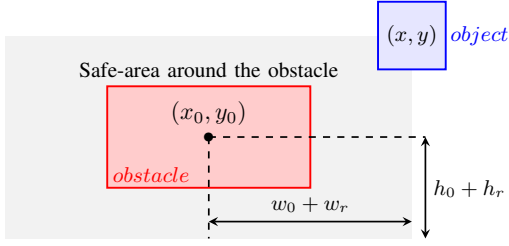
\begin{figure}[!t]
	\centering
	\scalebox{0.9}
	{
		\begin{tikzpicture}[node distance = 5em, auto, thick]
			\fill[gray!10!white] (0, 0) rectangle (6, 3);
			\fill[red!20!white, draw=red] (1.5, 0.75) rectangle (4.5, 2.25);
			\node (rect) at (3,2.5) [minimum width=2cm,minimum height=2cm] {\small{Safe-area around the obstacle}};
			\node [red] at (2.15, 0.95) {\small{$obstacle$}};
			
			\draw[dashed, line width=0.25mm] (3, 1.5) -- (6.2, 1.5);
			\draw[dashed, line width=0.25mm] (3, 1.5) -- (3, 0);
			
			\node [black] at (3,1.45) {\Large{\textbullet}};
			\node[] at (3, 1.85) {$(x_0, y_0)$};
			\node (rect) at (6,3) [draw=blue, fill=blue!10!white, minimum width=1cm,minimum height=1cm] {\small{$(x, y)$}};
			\node [blue] at (7, 3) {\small{$object$}};
			
			\draw[stealth-stealth, line width=0.25mm] (3, 0.25) -- (6, 0.25);
			\node[] at (4.5, 0.5) {\small{$w_0+w_r$}};
			\draw[stealth-stealth, line width=0.25mm] (6.2, 0) -- (6.2, 1.5);
			\node[rotate=0,anchor=north] at (6.85, 1) {\small{$h_0+h_r$}};
			
		\end{tikzpicture}
	}
	\caption{Hard constraint definition for the rectangle-shaped obstacles in 2D.}
	\label{fig:mpc-hard-constraint-rect}
\end{figure}

\begin{algorithm}[t!]
	\caption{MPC-HGG algorithm}\label{alg:exhgg_mpc}
	\small{
		\begin{algorithmic}[1]
			\State Given: RL pre-trained policy $\pi_{\theta}$, observation $s_0$, horizon $N$ and goal $g$
			\State $u_0 \gets (0,0,0)$
			\State $m \gets 1$
			\For{$t \gets 0$ to $T-1$}
			\State $p_t \gets getPos(s_t)$
			\State $v_t \gets getVel(s_t)$
			\State $a_t \gets \pi_{\theta}(s_t\:\vert\:g)$ \label{alg:line:rl}  \Comment{Get RL action}
			\State $g_t \gets p_t + a_t$  \Comment{Convert action to a new target}
			\If{$\lVert \mathbf{p_t, g} \rVert_2 \leq Nv_{\max}d_t$}  \Comment{Use MPC directly when goal is reachable}
			\State $g_t \gets g$ \label{alg:line:mpcgoal}
			\EndIf
			\State $z_1 \gets u_t \times p_t \times v_t$  \Comment{Setup first MPC state to start solver from}
			\State $p \gets \{g_t \times (o_k^i)_{i=1}^{N_o}\}_{k=1}^{N}$ \label{alg:line:mpcpar}  \Comment{Setup MPC parameters}
			\State $U_t \gets minimize \sum_{k=1}^{N} f_k(z_k, p_k)$ \label{alg:line:mpcprob}  \Comment{Solve MPC}
			\If{$U_t$ is feasible}  \Comment{A feasible solution found}
			\State $u_t \gets U_{t,1}$  \Comment{Take first MPC action}
			\State $a_t^{\prime} \gets getAction(u_t, a_t)$  \Comment{Convert MPC action to the MuJoCo action}
			\Else
			\State $m \gets m + 1$
			\If{$m < N$}
			\State $u_t \gets U_{t-1,m}$  \Comment{Try next prediction from the previous solution}
			\State $a_t^{\prime} \gets getAction(u_t, a_t)$ \label{alg:line:mpcprev}
			\Else
			\State $a_t^{\prime} \gets noAction(a_t)$ \label{alg:line:mpcdec}  \Comment{Decelerate robot}
			\EndIf
			\EndIf
			\State Perform $a_t^{\prime}$, get $s_{t+1}$ \label{alg:line:mpcres} \Comment{Perform action in the simulator and
				get a new observation}
			\EndFor
		\end{algorithmic}
	}
\end{algorithm}

\begin{figure*}[!t]
	\centering
	\begin{subfigure}[h]{0.325\textwidth}   
		\centering 
		\includegraphics[width=\textwidth]{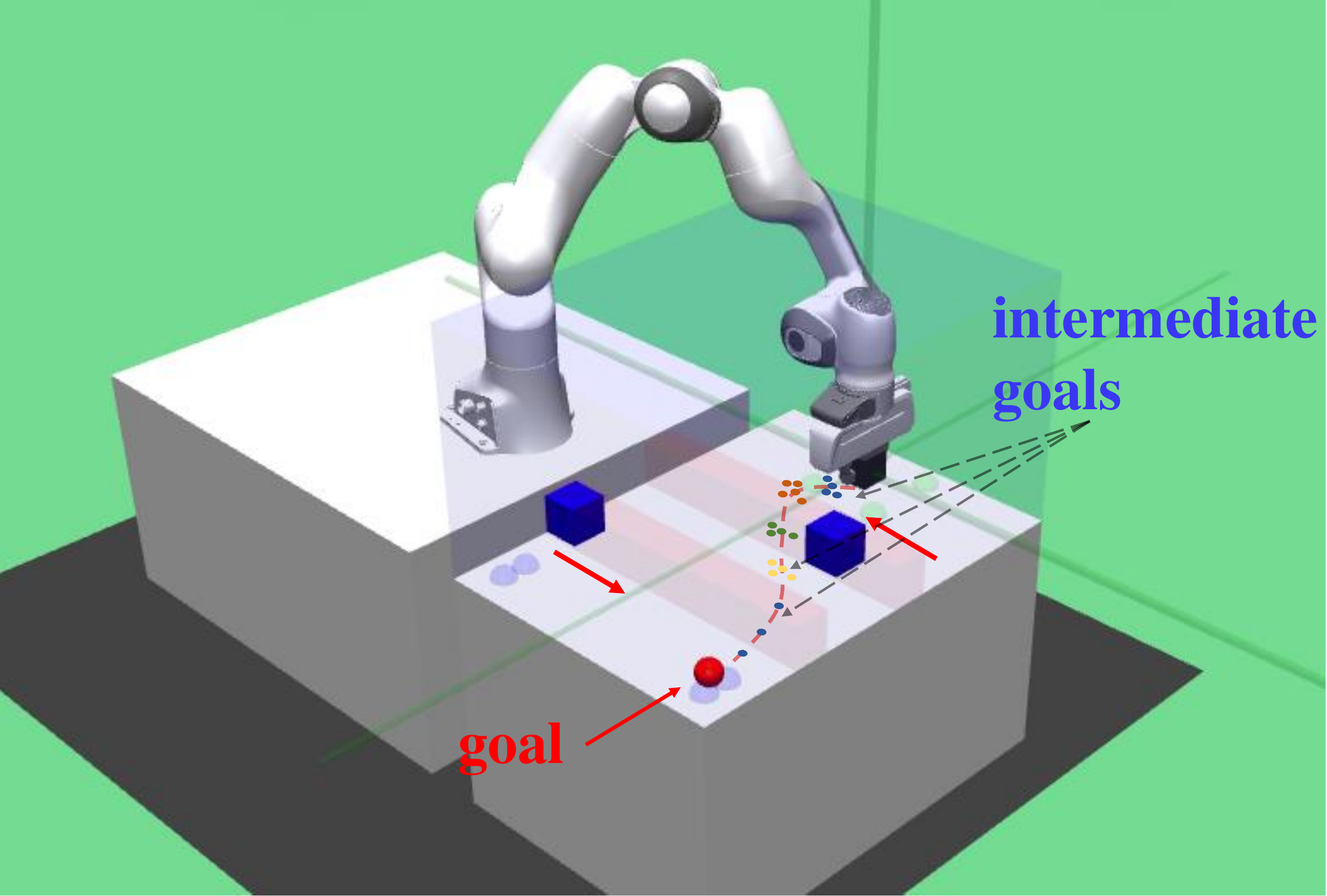}
		\caption{DynamicSquareObstacles.}     
	\end{subfigure}
	\begin{subfigure}[h]{0.217\textwidth}
		\centering
		\includegraphics[width=\textwidth]{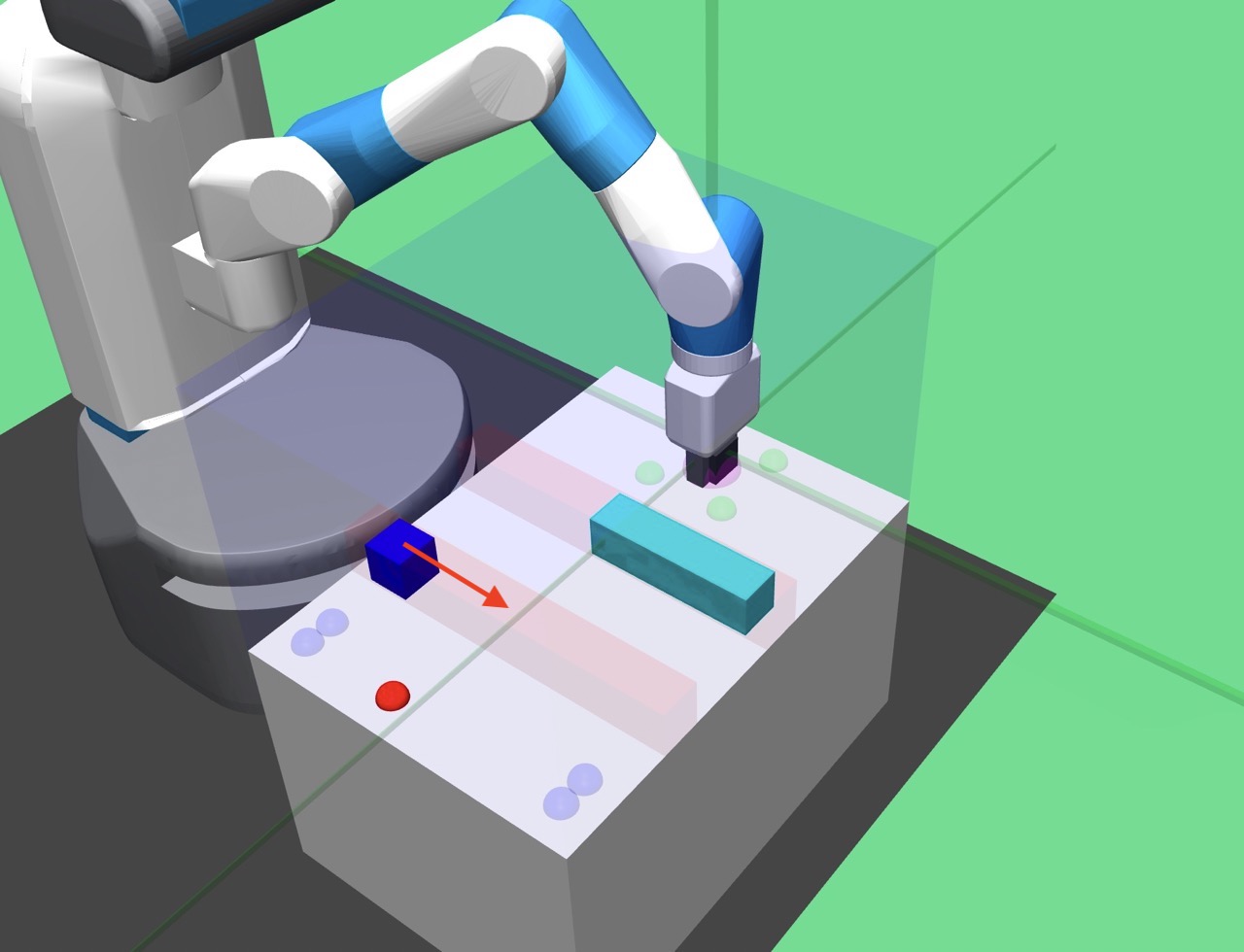}
		\caption{DynamicMixedObstacles. }   
	\end{subfigure}
	\begin{subfigure}[h]{0.217\textwidth}  
		\centering 
		\includegraphics[width=\textwidth]{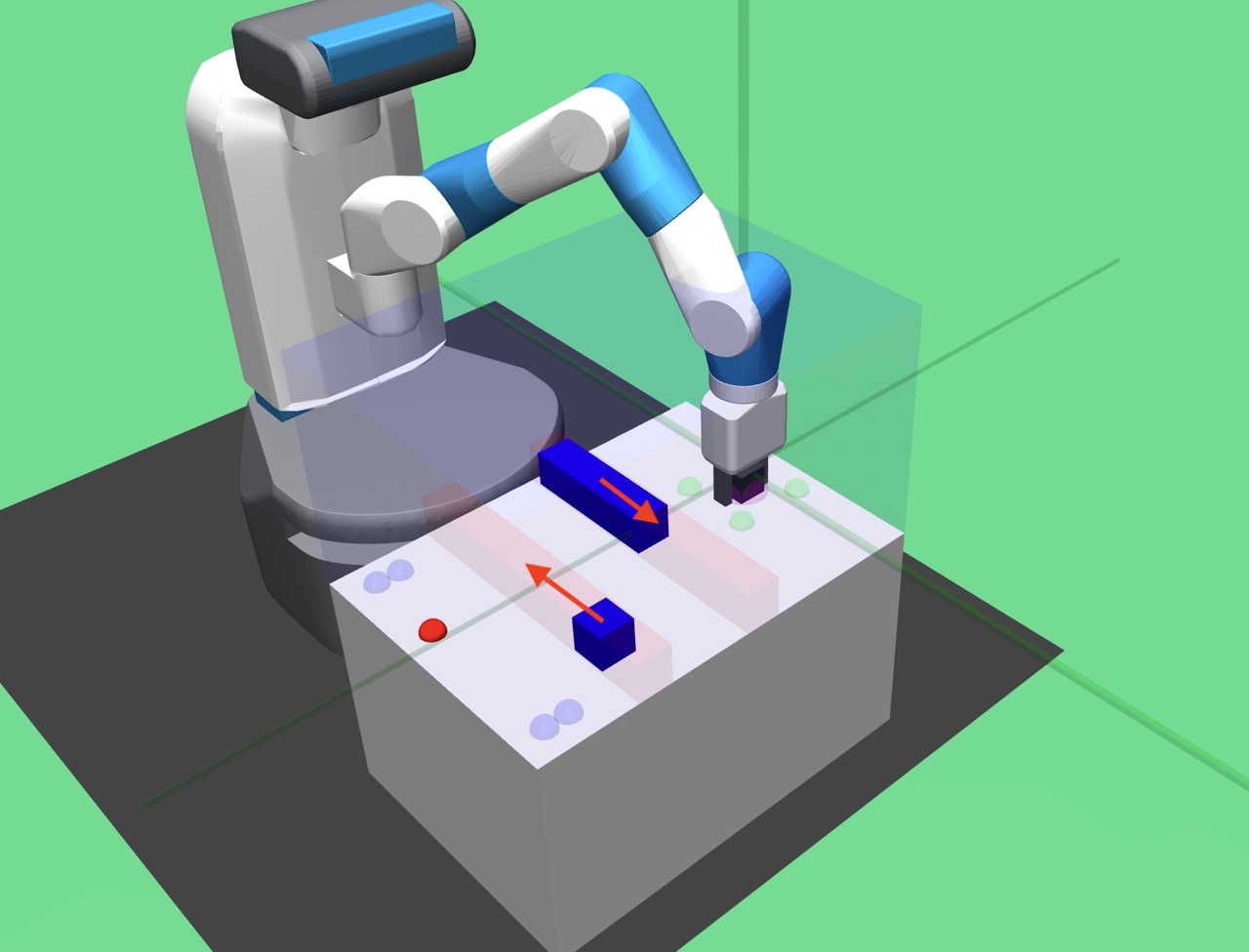}
		\caption{DynamicRecObstacles. }  
	\end{subfigure}
	\begin{subfigure}[h]{0.217\textwidth}   
		\centering 
		\includegraphics[width=\textwidth]{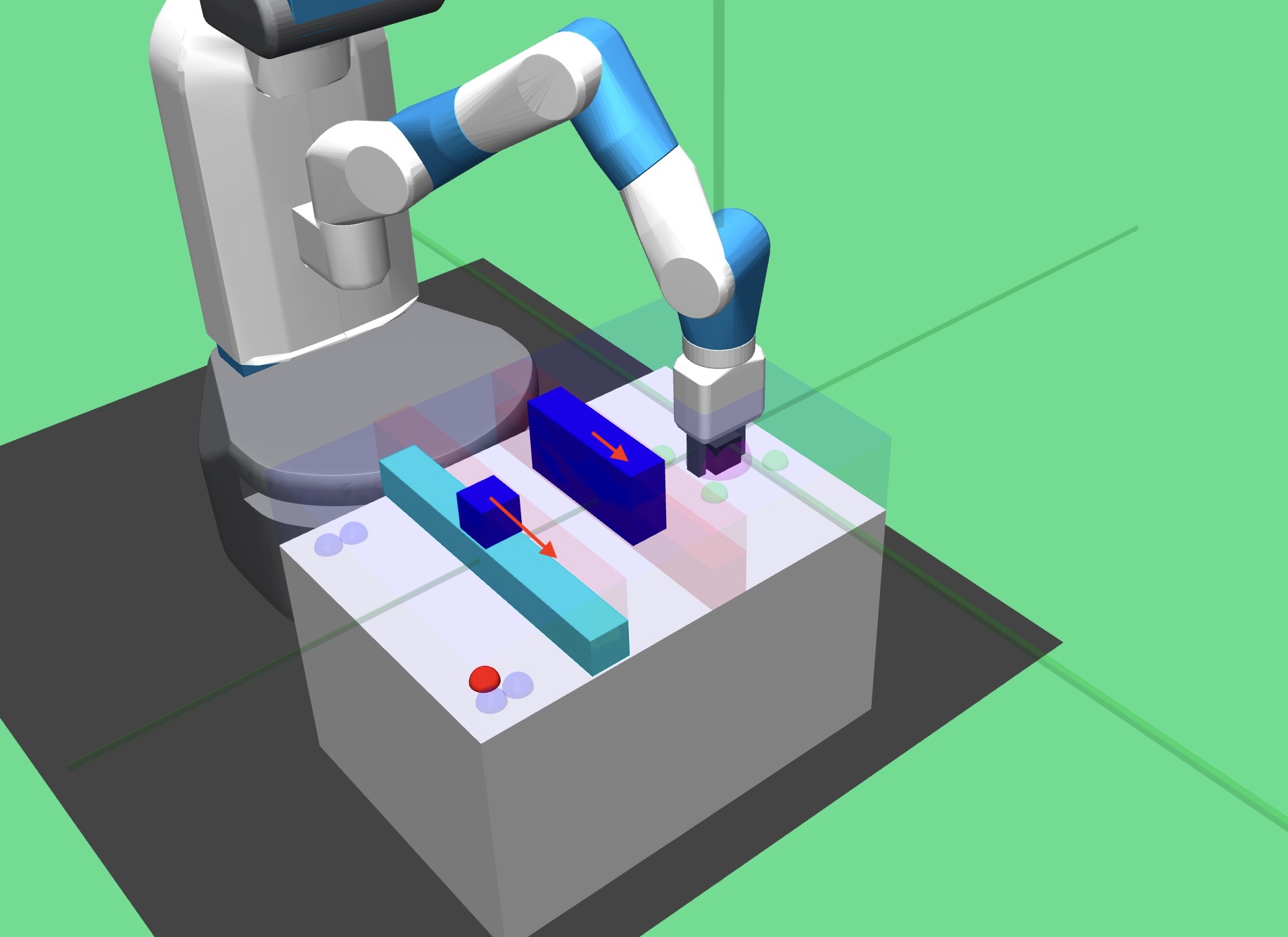}
		\caption{DynamicLiftedObstacles.}  
	\end{subfigure}
	
	\begin{subfigure}[h]{0.325\textwidth}   
		\centering 
		\includegraphics[width=\textwidth, trim={0cm 0 0cm 0},clip]{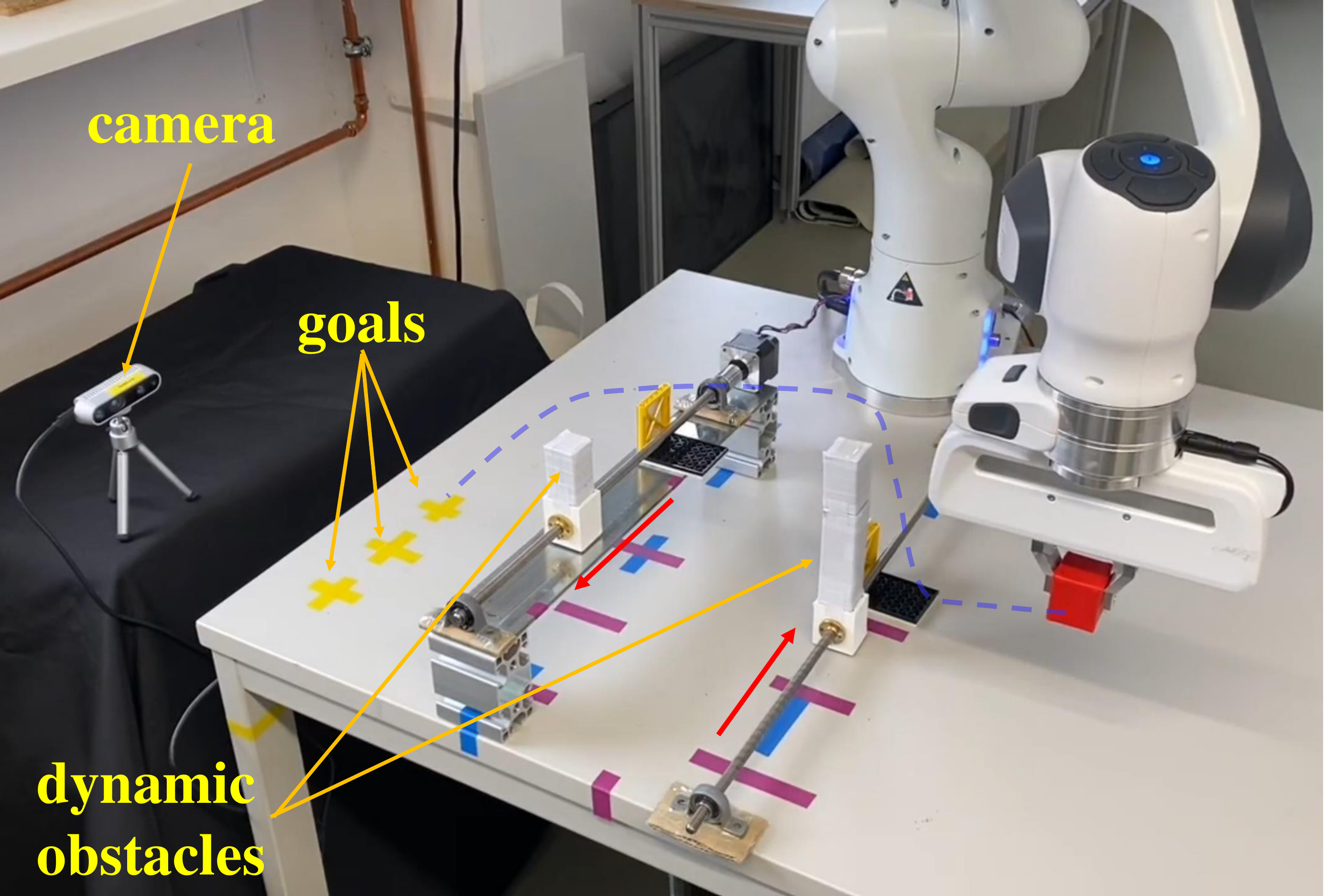}
		\caption{DynamicSquareObstacles.}  
	\end{subfigure}
	\begin{subfigure}[h]{0.217\textwidth}   
		\centering 
		\includegraphics[width=\textwidth, trim={0cm 0 0cm 0},clip]{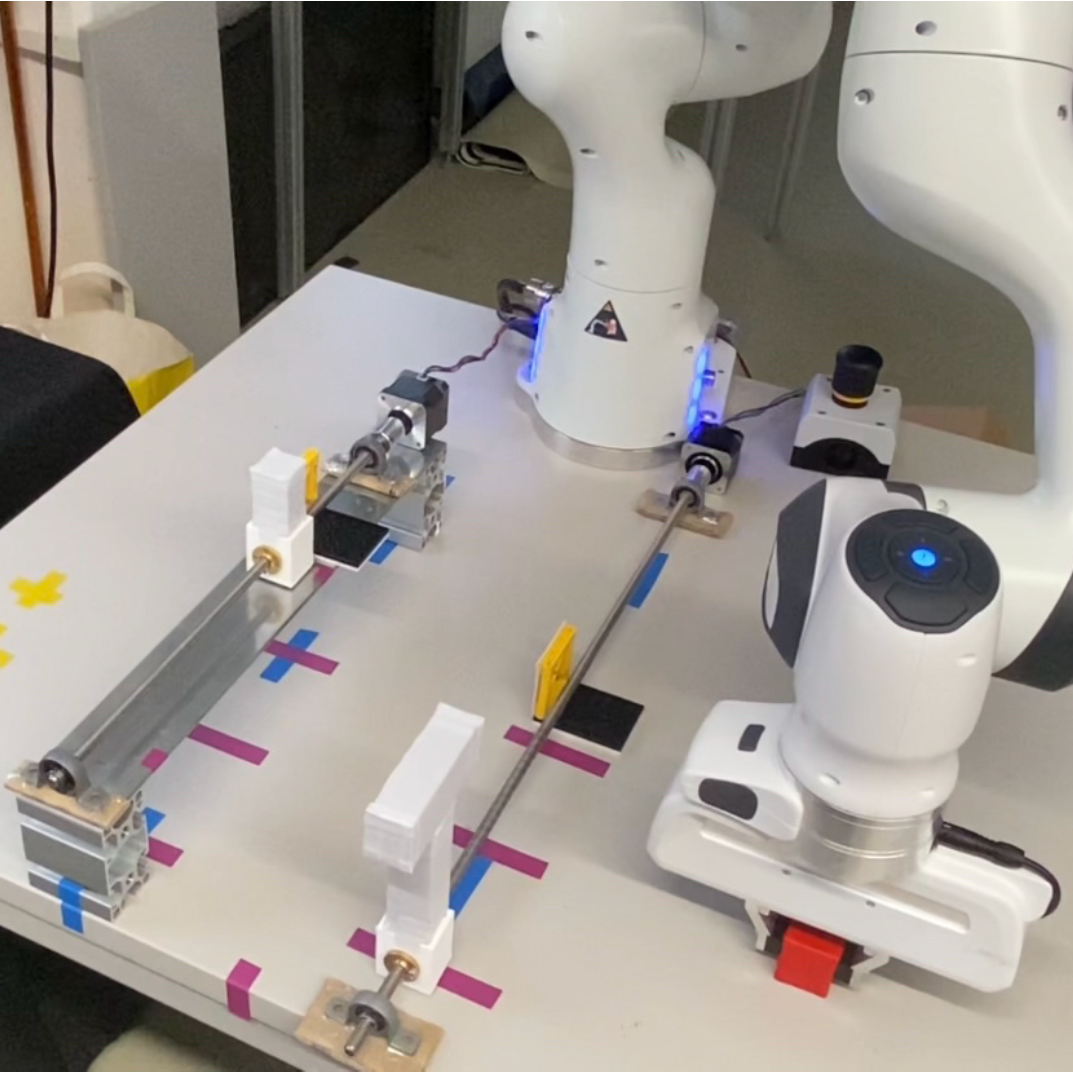}
		\caption{DynamicMixedObstacles.}  
	\end{subfigure}
	\begin{subfigure}[h]{0.217\textwidth}   
		\centering 
		\includegraphics[width=\textwidth, trim={0cm 0 0cm 0},clip]{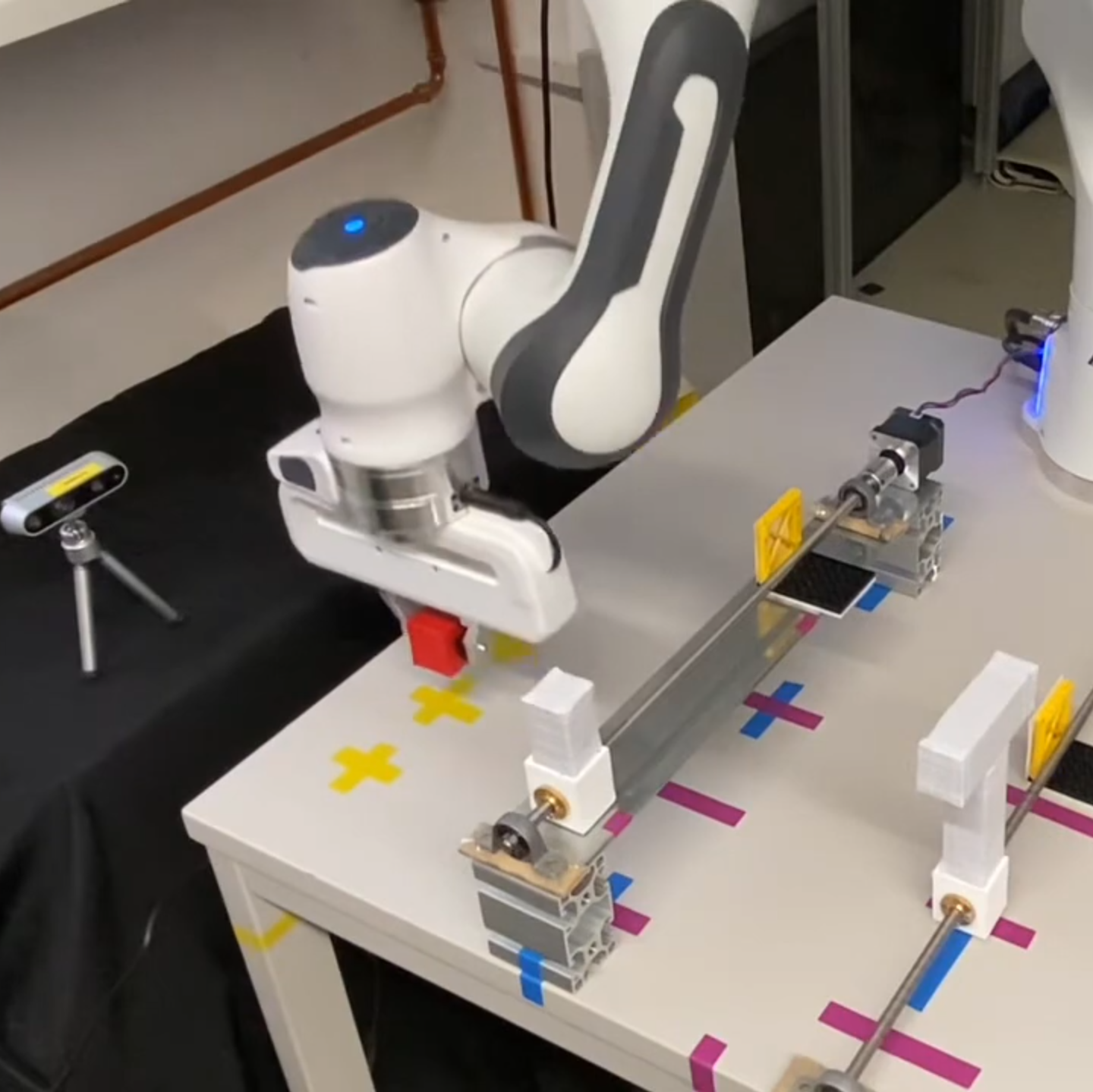}
		\caption{DynamicRecObstacles.}  
	\end{subfigure}
	\begin{subfigure}[h]{0.217\textwidth}   
		\centering 
		\includegraphics[width=\textwidth, trim={0cm 0 0cm 0},clip]{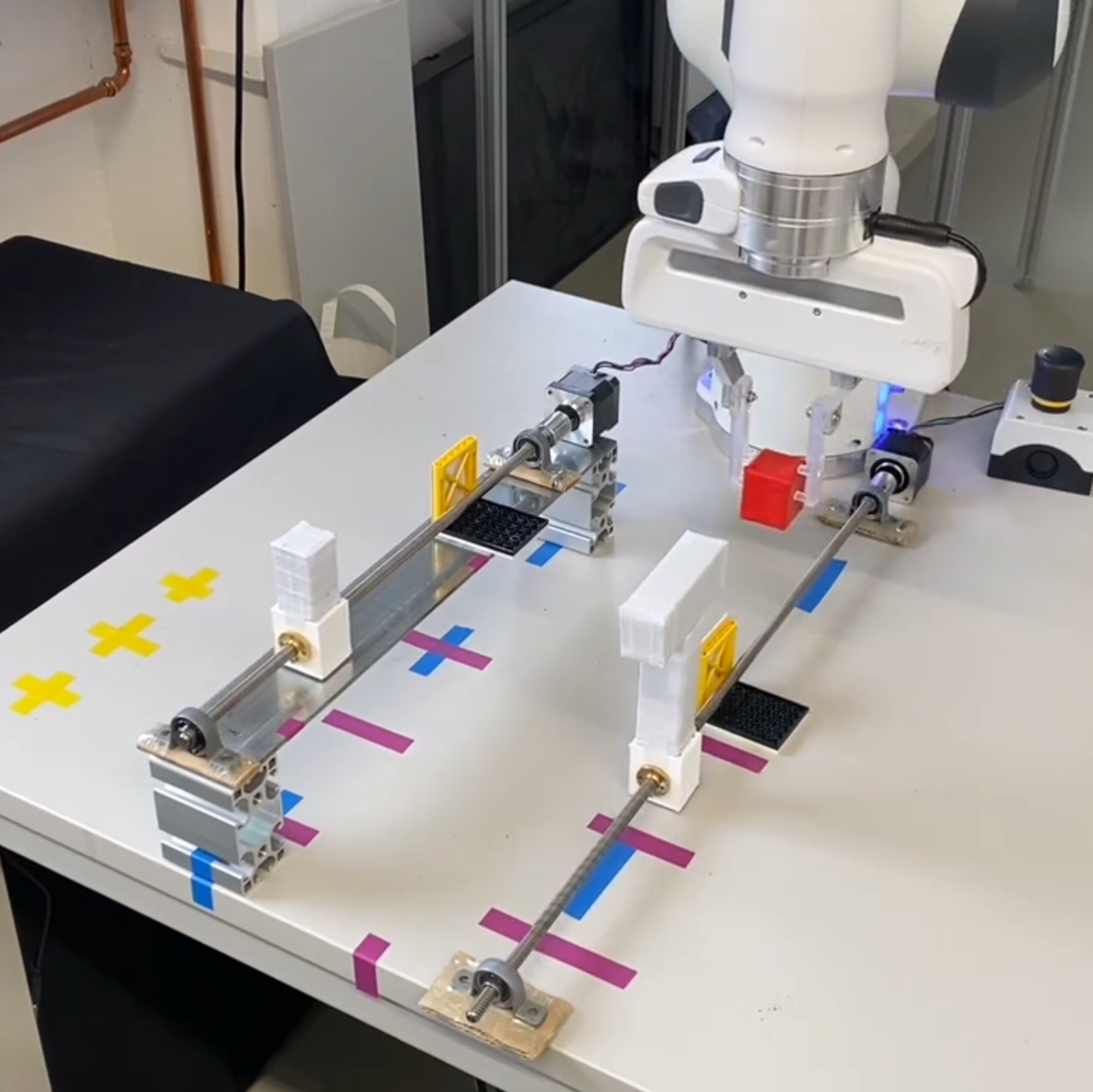}
		\caption{DynamicLiftedObstacles.}  
	\end{subfigure}
	
	\caption[Motor angles over time of a very well performing model.]{Robotic manipulation environments in the simulation and the real world.}
	\label{fig:scenarios}
\end{figure*}

\subsubsection{Cost function}\label{subsection:mpccostf}

The cost function $f_k(z_k, p_k)$ is formulated as follows:
\begin{equation}
	\label{eqn:mpccostfunc}
	\begin{aligned}\begin{gathered}
			f_k(z_k, p_k) = \left\{ \begin{array}{lll}
				w_1\lVert \mathbf{s, g} \rVert_2^2 + w_5 \xi^2 & \\ + w_2 F_x^2 + w_3 F_y^2  + w_4 F_z^2 & \mbox{for} & k < N  \\[0.5em]
				f_{k-1}(z_k, p_k) & \\ + w_6 v_x^2 + w_7 v_y^2 + w_8 v_z^2 & \mbox{for} & k = N
			\end{array}\right. \\
	\end{gathered}\end{aligned}
\end{equation}
, where we separate it to the stage cost and terminal cost.
The stage cost includes the Euclidean distance from the current
position $\bm{s}$ to the target goal $\bm{g}$ and penalization of the control commands to achieve the smooth trajectory.
For the terminal cost, the robot should reduce the velocity to zero when reaching the goal.
We denote $w_i$ as weights $\forall i \in \{1,\dots,8\}$ which should be fine-tuned to prioritize the penalties.

\subsubsection{Constraints}

The constraints for avoiding the obstacle is defined in the following form:
\begin{equation}
	\label{eqn:mpcconstrfunc_rect}
	\begin{aligned}\begin{gathered}
			h_{rect}(s_k, s_o, dim_o) = \frac{1}{2} max(\dfrac{|x-x_o|}{w_o + w_r}, \dfrac{|y-y_o|}{h_o + h_r},
			\dfrac{|z-z_o|}{d_o + d_r}) \\
			s_k = [x,y,z] \in \mathbb{R}^3\\
			s_o = [x_o,y_o,z_o] \in \mathcal{P}\\
			dim_o = [w_o,h_o,d_o] \in \mathcal{D}\\
	\end{gathered}\end{aligned}
\end{equation}
, where $w_r$, $h_r$, $d_r$ are width, height and depth of the rectangle describing the manipulated box together with
the gripper.
$w_0$, $h_0$, $d_0$ are width, height and depth of the bounding box of the obstacle.
Figure~\ref{fig:mpc-hard-constraint-rect} illustrates the proposed hard-constraint definition.
%
%
%
%
%
Since the $\max$ operator is not differentiable and thus not suitable for the gradient descent algorithms used in the MPC controller, we propose to approximate it using the smooth maximum~\cite{Lange2014ApplicationsOL}:
\begin{equation}
	\label{eqn:smoothmax}
	\begin{aligned}\begin{gathered}
			{\mathcal {S}}_{\alpha }(x_{1},\ldots ,x_{n})={\frac {\sum _{i=1}^{n}x_{i}e^{\alpha x_{i}}}{\sum _{i=1}^{n}e^{\alpha x_{i}}}}\\
			{\mathcal  {S}}_{\alpha }\to \max\:\mbox{as}\:\alpha \to \infty\\
	\end{gathered}\end{aligned}
\end{equation}

\subsection{MPC-HGG Algorithm}

The overall MPC-HGG algorithm is provided as Algorithm \ref{alg:exhgg_mpc}.
First, we extract the proposed RL action based on the current observation and convert it to a new intermediate goal for
the MPC in line~\ref{alg:line:rl}.
If the task can be solved within the planning horizon, we provide the main goal directly to the MPC
in line~\ref{alg:line:mpcgoal}.
Note that we still need RL action to control the gripper.
Once a sub-goal is selected by the RL \lstinline[columns=fixed]{planner}, the MPC can be formalized in
line~\ref{alg:line:mpcpar}.
We choose the horizon length $N$ for the MPC \lstinline[columns=fixed]{actor}, so that the robot is able to reach the proposed position within
the planning horizon $t_p$ and to stop there.
It is also important to choose the right integration time step $t_i < t_p$, so
that the robot will still be able to compute the next action within the time of the movement $t_i$.
Then we solve the MPC problem in line~\ref{alg:line:mpcprob}.
Finally, we perform the optimized action in the simulation (in line~\ref{alg:line:mpcres}) and receive a new
observation which is used in the next iteration.


\section{Experiments}

\newcommand{\drawEnvOne}{

        \begin{tikzpicture}[scale=0.5]
            \begin{axis}[
            legend style={
                         at={(0.5,1.15)},
                         anchor=north,
                         legend columns=-1,
                         /tikz/every even column/.append style={column sep=0.5cm},
                         font=\Large
                     },
            ybar=0pt,
            bar width=0.15,
            ymin=0, ymax=1.1,
            ytick={0, 0.2,...,1},
            xmin=-0.5, xmax=2.5,
            xtick={0, 1, 2},
            ylabel={Averaged Success Rate (\%)},
            xlabel={Tolerance of $N$ collision},
            label style={font=\Large},
            tick label style={font=\Large}  
            ]
            
            \addplot[draw=none, fill=blue!40, error bars/.cd,
            y dir=both,y explicit]
            coordinates {
                (0, 0.75) +- (0.04, 0.05)
                (1, 0.82) +- (0.04, 0.06)
                (2, 0.85) +- (0.04, 0.06)};
            \addplot[draw=none, fill=red!40, error bars/.cd,
            y dir=both,y explicit]
            coordinates {
                (0, 0.7) 
                (1, 0.82)
                (2, 0.83)};
            \addplot[draw=none, fill=green!30, error bars/.cd,
            y dir=both,y explicit]
            coordinates {
                (0, 1) +- (0.0, 0.0)
                (1, 1) +- (0.0, 0)
                (2, 1) +- (0.0, 0)};
            \legend{HGG, MPC, MPC-HGG}
            \end{axis}
        \end{tikzpicture}

}

\newcommand{\drawEnvTwo}{

        \begin{tikzpicture}[scale=0.5]
            \begin{axis}[
            legend style={
                         at={(0.5,1.15)},
                         anchor=north,
                         legend columns=-1,
                         /tikz/every even column/.append style={column sep=0.5cm},
                         font=\Large
                     },
            ybar=0pt,
            bar width=0.15,
            ymin=0, ymax=1.1,
            ytick={0, 0.2,...,1},
            xmin=-0.5, xmax=2.5,
            xtick={0, 1, 2},
            ylabel={Averaged Success Rate (\%)},
            xlabel={Tolerance of $N$ collision},
            label style={font=\Large},
            tick label style={font=\Large}  
            ]
            
            \addplot[draw=none, fill=blue!40, error bars/.cd,
            y dir=both,y explicit]
            coordinates {
                (0, 0.35) +- (0.04, 0.05)
                (1, 0.82) +- (0.04, 0.06)
                (2, 0.9)  +- (0.02, 0.02)};
            \addplot[draw=none, fill=red!40, error bars/.cd,
            y dir=both,y explicit]
            coordinates {
                (0, 0.3) 
                (1, 0.3)
                (2, 0.3)};
            \addplot[draw=none, fill=green!30, error bars/.cd,
            y dir=both,y explicit]
            coordinates {
                (0, 1) +- (0.0, 0.0)
                (1, 1) +- (0.0, 0)
                (2, 1) +- (0.0, 0)};
            \legend{HGG, MPC, MPC-HGG}
            \end{axis}
        \end{tikzpicture}

}

\newcommand{\drawEnvThree}{

        \begin{tikzpicture}[scale=0.5]
            \begin{axis}[
            legend style={
                         at={(0.5,1.15)},
                         anchor=north,
                         legend columns=-1,
                         /tikz/every even column/.append style={column sep=0.5cm},
                         font=\Large
                     },
            ybar=0pt,
            bar width=0.15,
            ymin=0, ymax=1.1,
            ytick={0, 0.2,...,1},
            xmin=-0.5, xmax=2.5,
            xtick={0, 1, 2},
            ylabel={Averaged Success Rate (\%)},
            xlabel={Tolerance of $N$ collision},
            label style={font=\Large},
            tick label style={font=\Large}  
            ]
            
            \addplot[draw=none, fill=blue!40, error bars/.cd,
            y dir=both,y explicit]
            coordinates {
                (0, 0.3) +- (0.04, 0.05)
                (1, 0.6) +- (0.04, 0.06)
                (2, 0.8)  +- (0.02, 0.02)};
            \addplot[draw=none, fill=red!40, error bars/.cd,
            y dir=both,y explicit]
            coordinates {
                (0, 0.82) 
                (1, 0.9)
                (2, 0.92)};
            \addplot[draw=none, fill=green!30, error bars/.cd,
            y dir=both,y explicit]
            coordinates {
                (0, 1) +- (0.0, 0.0)
                (1, 1) +- (0.0, 0)
                (2, 1) +- (0.0, 0)};
            \legend{HGG, MPC, MPC-HGG}
            \end{axis}
        \end{tikzpicture}

}

\newcommand{\drawEnvFour}{

        \begin{tikzpicture}[scale=0.5]
            \begin{axis}[
            legend style={
                         at={(0.5,1.15)},
                         anchor=north,
                         legend columns=-1,
                         /tikz/every even column/.append style={column sep=0.5cm},
                         font=\Large
                     },
            ybar=0pt,
            bar width=0.15,
            ymin=0, ymax=1.1,
            ytick={0, 0.2,...,1},
            xmin=-0.5, xmax=2.5,
            xtick={0, 1, 2},
            ylabel={Averaged Success Rate (\%)},
            xlabel={Tolerance of $N$ collision},
            label style={font=\Large},
            tick label style={font=\Large}  
            ]
            
            \addplot[draw=none, fill=blue!40, error bars/.cd,
            y dir=both,y explicit]
            coordinates {
                (0, 0.3) +- (0.04, 0.05)
                (1, 0.6) +- (0.04, 0.06)
                (2, 0.8)  +- (0.02, 0.02)};
            \addplot[draw=none, fill=red!40, error bars/.cd,
            y dir=both,y explicit]
            coordinates {
                (0, 0.) 
                (1, 0.)
                (2, 0)};
            \addplot[draw=none, fill=green!30, error bars/.cd,
            y dir=both,y explicit]
            coordinates {
                (0, 1) +- (0.0, 0.0)
                (1, 1) +- (0.0, 0)
                (2, 1) +- (0.0, 0)};
            \legend{HGG, MPC, MPC-HGG}
            \end{axis}
        \end{tikzpicture}

}
\begin{figure*}[t!]
	\centering
	\begin{subfigure}[t]{.22\textwidth}
		\centering
		\drawEnvOne
		\caption{DynamicSquareObstacles.}
		\label{subfig:fig_KukaPickNoObstacle_eta}
	\end{subfigure}
	\hspace{0.4cm}
	\begin{subfigure}[t]{.22\textwidth}
		\centering
		\drawEnvTwo
		\caption{DynamicMixedObstacles.}
		\label{subfig:fig_KukaPickNoObstacle_eta}
	\end{subfigure}
	\hspace{0.4cm}
	\begin{subfigure}[t]{.22\textwidth}
		\centering
		\drawEnvThree
		\caption{DynamicRecObstacles.}
		\label{subfig:fig_KukaPickNoObstacle_eta}
	\end{subfigure}
	\hspace{0.4cm}
	\begin{subfigure}[t]{.22\textwidth}
		\centering
		\drawEnvFour
		\caption{DynamicLiftedObstacles.}
		\label{subfig:fig_KukaPickNoObstacle_eta}
	\end{subfigure}
	\caption{Success rates of the collision avoidance testing of MPC-HGG, HGG, and the MPC controller.}
	\label{fig:success_rate}
\end{figure*}
\begin{figure*}[!tbp]
	\centering
	\subfloat[step8][Step 12]{\includegraphics[height=0.38\linewidth]
		{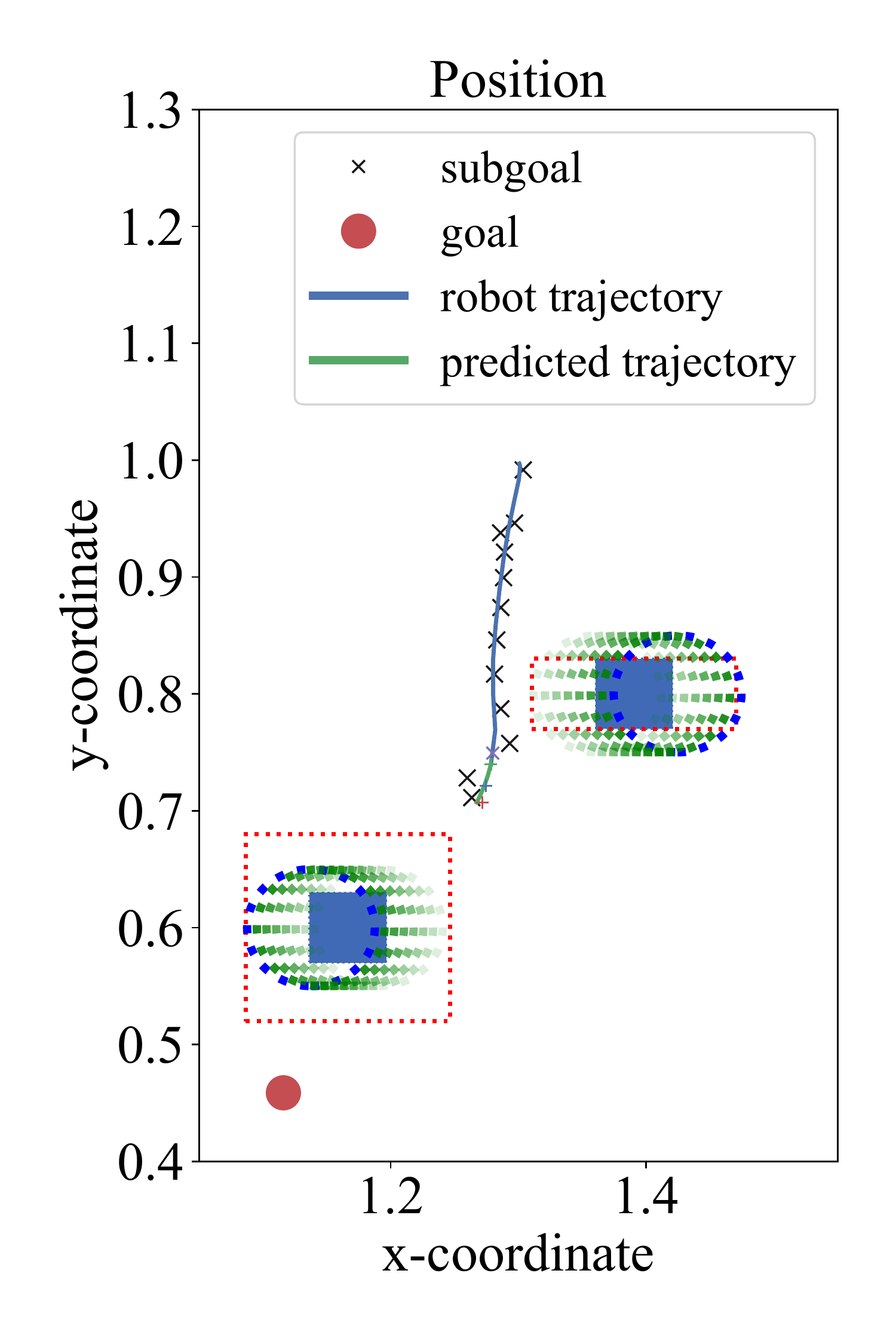}\label{fig:exp_sqr_obstacle:a}}
	\subfloat[step11][Step 15]{\includegraphics[height=0.38\linewidth]
		{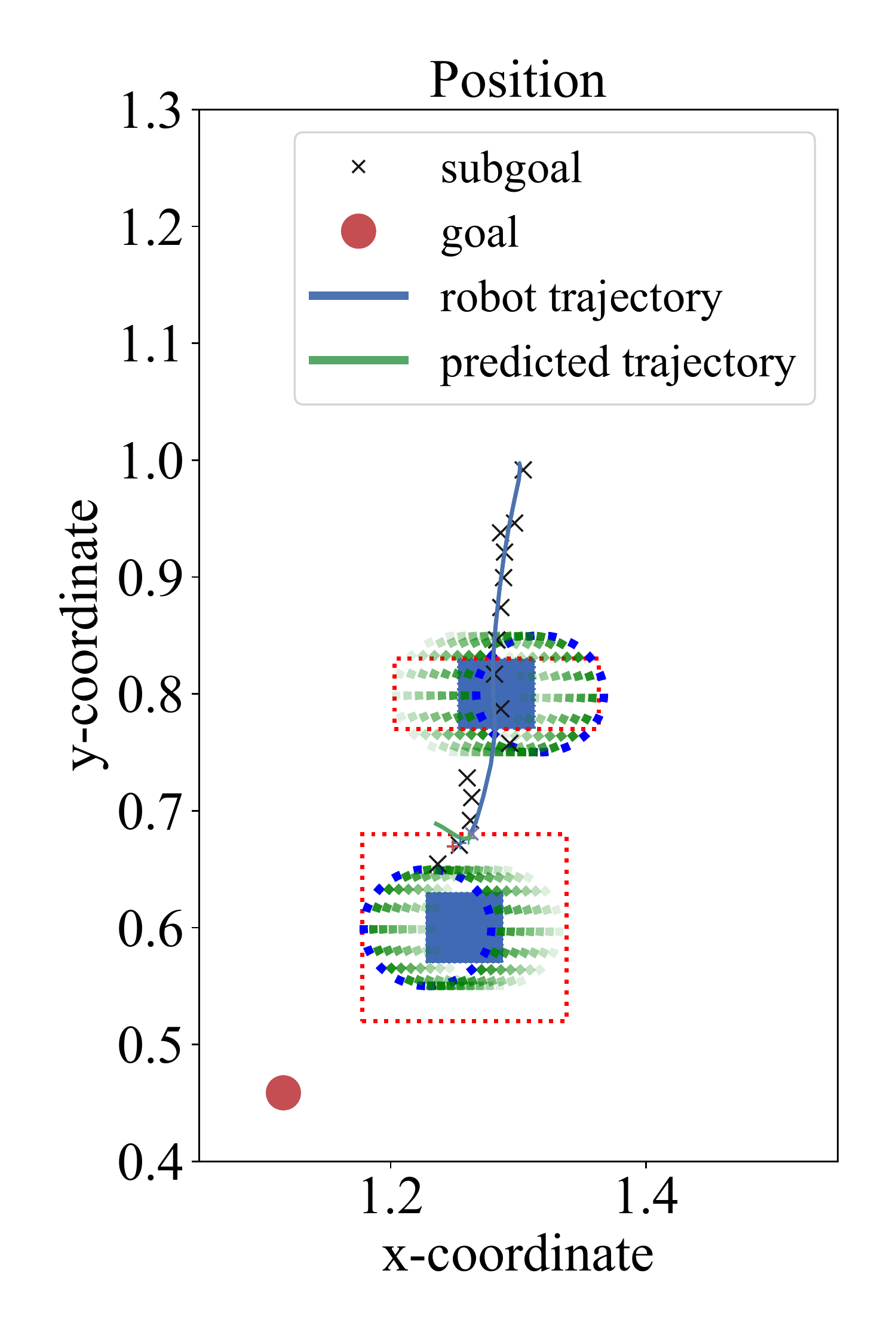}\label{fig:exp_sqr_obstacle:b}}
	\subfloat[step26][Step 19]{\includegraphics[height=0.38\linewidth]
		{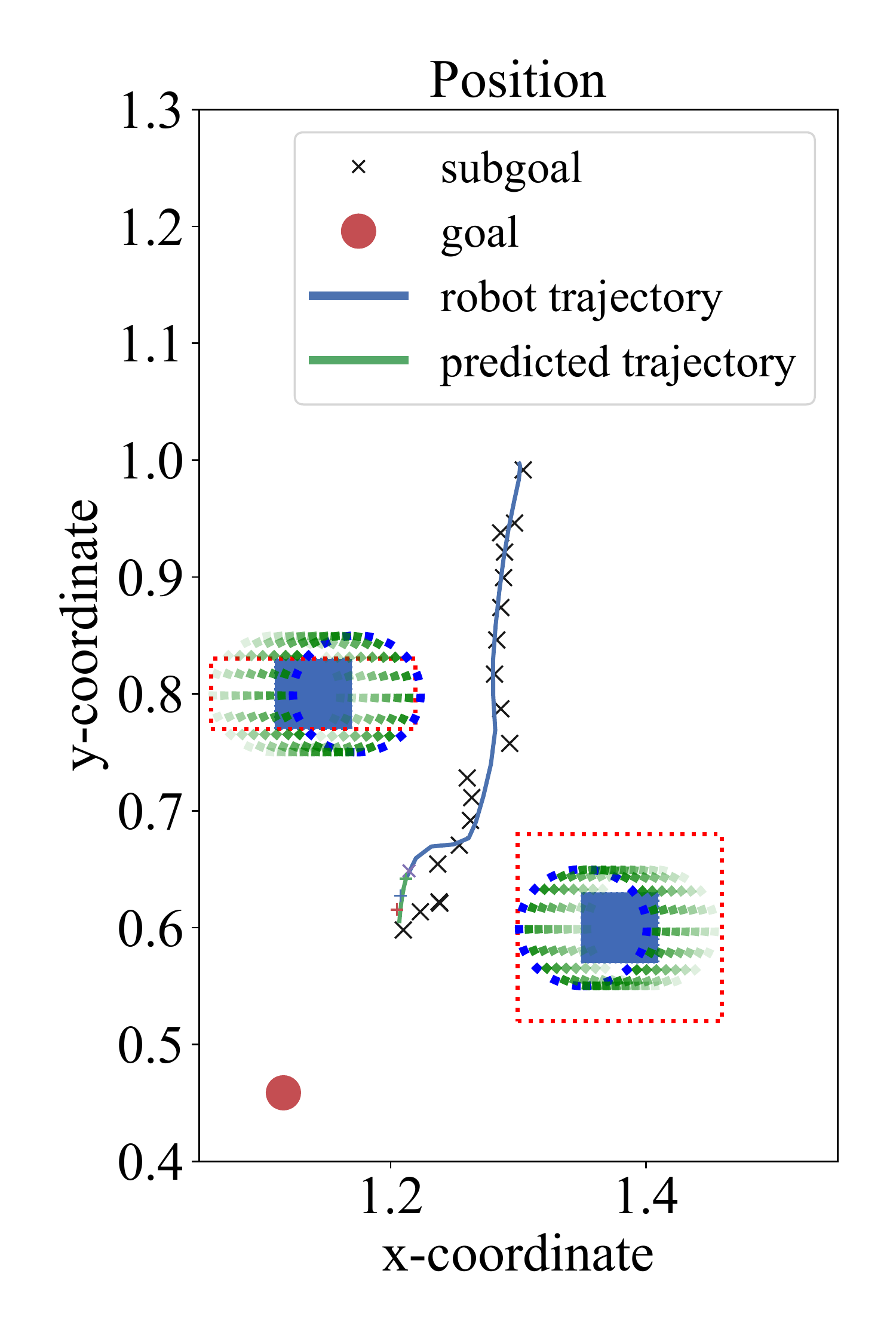}\label{fig:exp_sqr_obstacle:c}}
	\subfloat[step47][Step 25]{\includegraphics[height=0.38\linewidth]
		{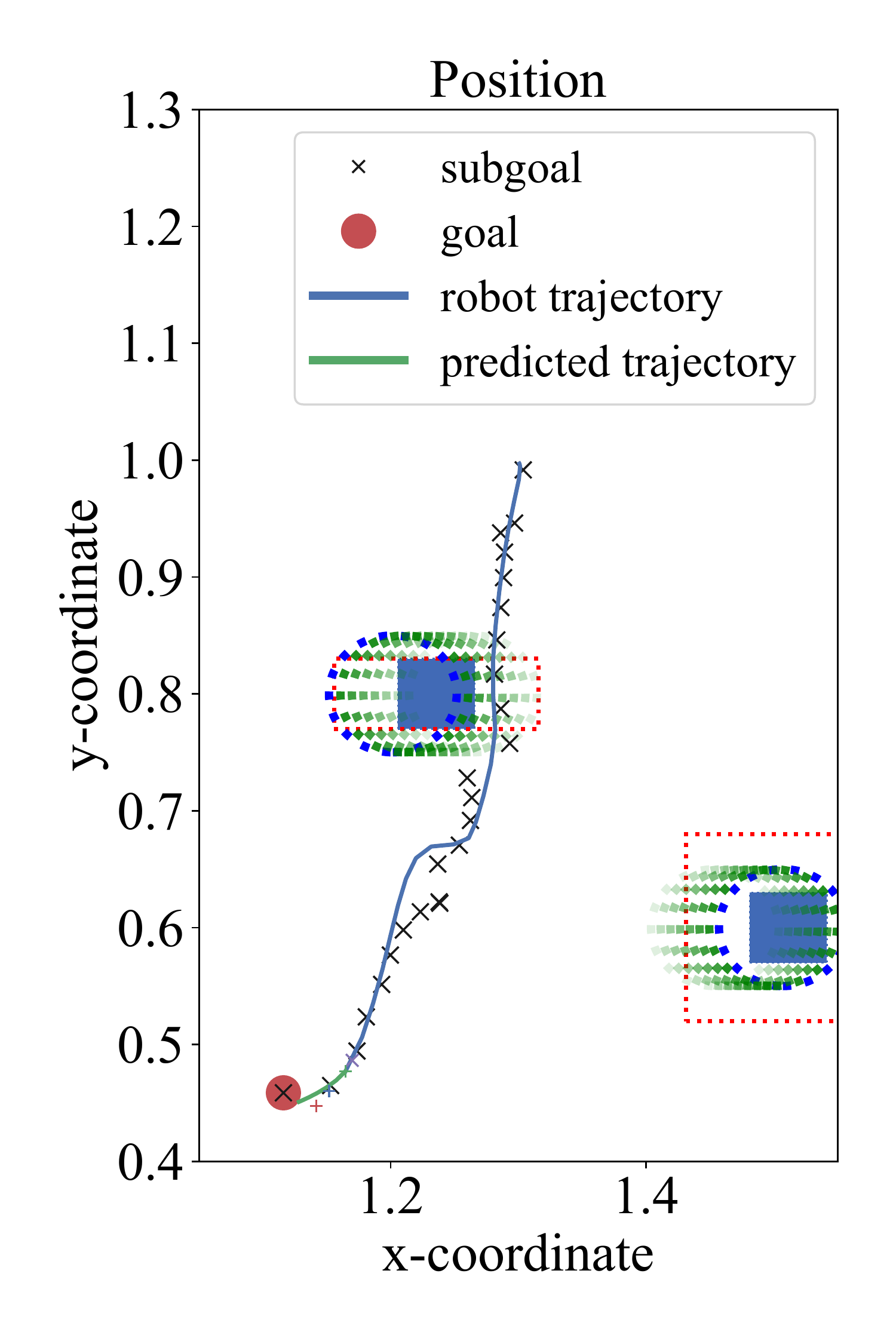}\label{fig:exp_sqr_obstacle:d}}
	\caption{Example of a resolved collision in the \textit{DynamicSquareObstacles} environment.
		The red dot represents the goal position and the cross marks are the intermediate goals suggested by the RL \lstinline[columns=fixed]{planner}. The dashed blue circles are the current position of the obstacle and the dashed green circles are the predicted states in the near future. }
	\label{fig:exp_sqr_obstacle}
\end{figure*}

In this section, we present an empirical evaluation of the performance of MPC-HGG in comparison to HGG and vanilla MPC across four distinct MuJoCo environments.
These environments are variants of the Fetch gripper environments introduced by \cite{plappert2018multi}, and are characterized by the presence of long-horizon goal-reaching tasks with static or dynamic obstacles.  
\subsection{Simulations}
All our tasks are simulated in MuJoCo \cite{todorov2012mujoco} and are performed in the real world, in which a Panda robot with a gripper is controlled to push a puck through environments with dynamic obstacles (see Figure \ref{fig:scenarios}).
These tasks share the following characteristics.
First, the agent receives a state containing the joint positions and velocities of the robotic arm. 
This information is directly retrieved from the simulation or from the robotic arm. 
Second, the robot is controlled by a three-dimensional vector describing the end effector's position. 
In the case of enabled gripper control, the gripper’s opening control parameter is added as the fourth component.
Third, the accessible goal space $\mathcal{G}_A$ is defined by a 2D region on the table. 
Fourth, we obtained the positional information of obstacles directly from the simulation. Conversely, in the physical experiment, a side-view camera was employed to capture the relevant positional data. Notably, we determined the distance threshold required to ascertain the successful attainment of a goal to be $0.05$ m.

The difficulty of the four tasks is gradually increased, and a brief description of each environment is given as follows.
\begin{enumerate}
	\item \textit{DynamicSquareObstacles} (see Figure \ref{fig:scenarios}(a): In this environment, there are two dynamic square obstacles that can move linearly in two directions with a velocity randomly sampled between $0.6$ m/s and $0.9$ m/s.
	The direction of the movement is also randomized.
	The transparent red regions represent the area that can be blocked by the obstacles.
	\item \textit{DynamicMixedObstacles} (see Figure \ref{fig:scenarios}(b): In this environment, there is one dynamic square obstacle and one static big rectangle obstacle.
	Both obstacles are sampled randomly inside the red region.
	The square obstacle is moving with a velocity sampled between $0.6$ m/s and $0.9$ m/s.
	\item \textit{DynamicRectObstacles} (see Figure \ref{fig:scenarios}(c): In this environment, the speed of the rectangle obstacle is chosen randomly from the interval between $0.2$m/s and $0.6$m/s.
	\item \textit{DynamicLiftedObstacles} (see Figure \ref{fig:scenarios}(d): In this environment, a rectangular static obstacle is placed under one of the dynamic obstacles. 
	Therefore, the robot must lift the object above the static obstacle and subsequently lower it down to the goal positions. 
	Dynamic obstacles are sampled randomly inside the red region.
\end{enumerate}



Figure \ref{fig:success_rate} shows the testing success rate of HGG, MPC, and MPC-HGG, which is calculated by averaging the performance of the best policy from each algorithm in $100$ episodes.
These tests have a tolerance parameter $N \in \{0, 1, 2\} $ for the number of collisions that can be allowed per episode. 
If the number of collisions surpasses $N$, then the episode is terminated as a failure.
The most remarkable results can be observed, in all four environments; MPC-HGG can learn safe obstacle-avoiding behavior with a success rate of $100\%$ across different numbers of tolerance parameters, while the other algorithms are not able to solve tasks without collisions.
In the \textit{DynamicSquareObstacles} environment (see Figure \ref{fig:success_rate} (a), the HGG controller achieves a success rate around $75\%$ when $N=0$ and slightly increases its performance when $N=1$ and $N=2$.
Since the MPC controller is a deterministic method, it demonstrates a consistently stable performance without any associated error bars.
The proposed MPC-HGG controller combines the advantage of long-horizon planning of the HGG controller with the short-horizon safety guarantee of the MPC controller, leading to a highly effective solution with a remarkable success rate of $100\%$.
Similar results can also observed in the other three scenarios (see Figure \ref{fig:success_rate} (b,c,d)).

Figure \ref{fig:exp_sqr_obstacle} shows a successful obstacle avoiding behavior in the \textit{DynamicSquareObstacles} environment.
At step $15$, the subgoals proposed by the RL \lstinline[columns=fixed]{planner} (marked by ```$\times$") lead to an intersection between the puck and the second dynamic obstacle.
In the effect of the MPC \lstinline[columns=fixed]{actor}, the agent predicts the upcoming of the obstacle and calculates a safe path by shifting its moving direction away from the obstacle (see Figure \ref{fig:exp_sqr_obstacle:c}, step 19).
At step $25$, the puck is successfully placed to the goal position (see Figure \ref{fig:exp_sqr_obstacle:d}).
The simulation videos show the obstacle-avoidance movements in detail \footnote{\url{https://videoviewsite.wixsite.com/mpc-hgg}}.





\subsection{Real-world Experiments}



As illustrated in Figure \ref{fig:scenarios}(e), the real-world experiment setup uses a Franka Emika Panda robotic arm and an Intel RealSense camera to obtain the coordinates of the manipulatable object. 
Two leading screws that carry white blocks are controlled by stepper motors and act as the dynamic obstacles.
The control rate is consistent with the simulation, which is $20$ Hz across all experiments.
The ForcesPro \cite{FORCESPro} is used as the MPC solver for fast-speed computation.
The overall computational time is less than $3$ ms for a planning horizon of $8$ steps.

As in the four environments in the simulation, we also create four environments featuring the Panda robotic arm with the obstacles (see Figure \ref{fig:scenarios}).
It should be noted that unlike the simulation that obtains the coordinates of the obstacles directly, we use a camera to gather this information by tracking the ArUco Maker attached to moving obstacles in the real world.

We took the policy directly from each of the tasks trained in the simulation and deployed it in the real world without any fine tuning. 
Inspired by the experiment performed by HER \cite{her}, we also add Gaussian noise to the observed object’s position during policy training to compensate for the small errors introduced by
the camera, which can increase the success rate of these tasks.
The performances of these four tasks demonstrate that the policy can be successfully transferred to the corresponding tasks in the real world.
In each scenario, the task is performed three times by selecting different goal locations.
As shown in the video, the robot arm can always successfully approach the target position in all trajectories while actively avoiding the dynamic obstacles. 
Once the MPC \lstinline[columns=fixed]{actor} anticipates a collision, it will control the robotic arm to move away from the moving direction of the obstacle and replan its trajectory to the target goal position.
The real-world experiment videos can be found at the project's website.
As shown in the video, the robot arm can always successfully
avoid the obstacles and approach the target position in every trial.

\begin{figure}
	\centering
	\includegraphics[width=0.48\textwidth]{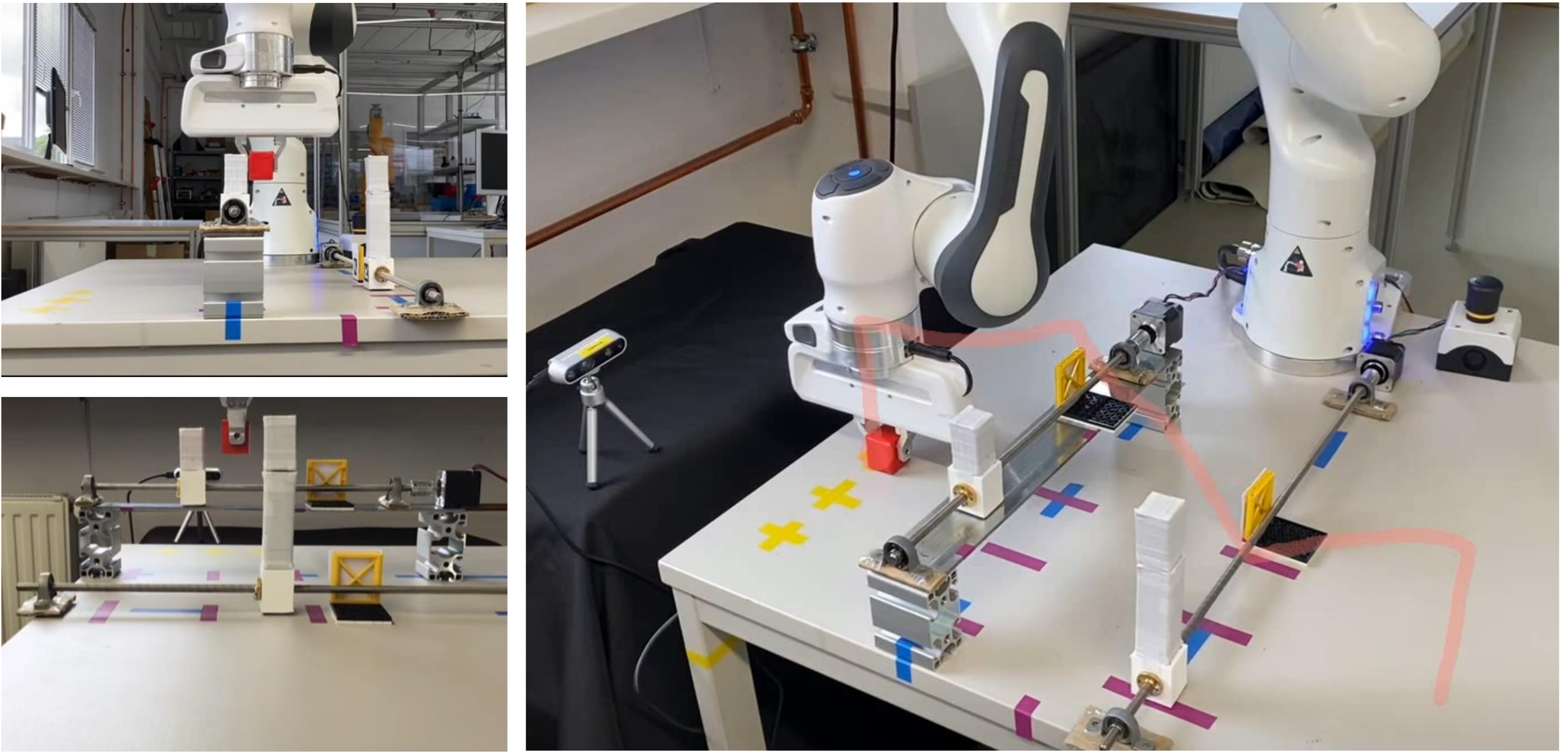}
	\caption{The scenario scene from the front and right (left).
		Right: One example illustrates the trajectory of the robotic arm in the \textit{DynamicSquareObstacles} environment from the isometric view (right).
		This animated exemplary video is available at the project's website.}
	\label{fig:my_label}
\end{figure}


\section{Conclusion}
In this paper, we have introduced a safety guaranteed planning framework for manipulation tasks based on a RL \lstinline[columns=fixed]{planner} and a MPC \lstinline[columns=fixed]{actor}.
We showed that our RL planner exhibits the capability to choose a sequence of intermediary goals that culminate in achieving the ultimate targets, while proficiently circumventing dynamic impediments encountered in its trajectory.
We also demonstrated that our MPC \lstinline[columns=fixed]{actor} is able to accomplish the designated goals proposed the RL \lstinline[columns=fixed]{planner} while ensuring no collision with dynamic obstacles.
For future work, we will explore more advanced methods to ensure the safety of the manipulation operation.

\section*{Acknowledgment}

This project/research has received funding from the European
Union’s Horizon 2020 Framework Programme for Research and
Innovation under the Specific Grant Agreement No.945539 (Human Brain Project SGA3).
The authors also acknowledge the financial support by the Bavarian State Ministry for Economic Affairs, Regional Development and Energy (StMWi) for the Lighthouse Initiative KI.FABRIK, (Phase 1: Infrastructure as well as the research and development program under, grant no. DIK0249).


%








\bibliographystyle{IEEEtran}
\bibliography{bibliography}

\begin{thebibliography}{10}
\providecommand{\url}[1]{#1}
\csname url@samestyle\endcsname
\providecommand{\newblock}{\relax}
\providecommand{\bibinfo}[2]{#2}
\providecommand{\BIBentrySTDinterwordspacing}{\spaceskip=0pt\relax}
\providecommand{\BIBentryALTinterwordstretchfactor}{4}
\providecommand{\BIBentryALTinterwordspacing}{\spaceskip=\fontdimen2\font plus
\BIBentryALTinterwordstretchfactor\fontdimen3\font minus
  \fontdimen4\font\relax}
\providecommand{\BIBforeignlanguage}[2]{{%
\expandafter\ifx\csname l@#1\endcsname\relax
\typeout{** WARNING: IEEEtran.bst: No hyphenation pattern has been}%
\typeout{** loaded for the language `#1'. Using the pattern for}%
\typeout{** the default language instead.}%
\else
\language=\csname l@#1\endcsname
\fi
#2}}
\providecommand{\BIBdecl}{\relax}
\BIBdecl

\bibitem{9466373}
Z.~Bing, M.~Brucker, F.~O. Morin, R.~Li, X.~Su, K.~Huang, and A.~Knoll,
  ``Complex robotic manipulation via graph-based hindsight goal generation,''
  \emph{IEEE Transactions on Neural Networks and Learning Systems}, vol.~33,
  no.~12, pp. 7863--7876, 2022.

\bibitem{hwangbo2019learning}
J.~Hwangbo, J.~Lee, A.~Dosovitskiy, D.~Bellicoso, V.~Tsounis, V.~Koltun, and
  M.~Hutter, ``Learning agile and dynamic motor skills for legged robots,''
  \emph{Science Robotics}, vol.~4, no.~26, p. eaau5872, 2019.

\bibitem{kollar2008trajectory}
T.~Kollar and N.~Roy, ``Trajectory optimization using reinforcement learning
  for map exploration,'' \emph{The International Journal of Robotics Research},
  vol.~27, no.~2, pp. 175--196, 2008.

\bibitem{her}
\BIBentryALTinterwordspacing
M.~Andrychowicz, F.~Wolski, A.~Ray, J.~Schneider, R.~Fong, P.~Welinder,
  B.~McGrew, J.~Tobin, P.~Abbeel, and W.~Zaremba, ``Hindsight experience
  replay,'' 2017. [Online]. Available: \url{https://arxiv.org/abs/1707.01495}
\BIBentrySTDinterwordspacing

\bibitem{hggalg}
\BIBentryALTinterwordspacing
Z.~Ren, K.~Dong, Y.~Zhou, Q.~Liu, and J.~Peng, ``Exploration via hindsight goal
  generation,'' 2019. [Online]. Available:
  \url{https://arxiv.org/abs/1906.04279}
\BIBentrySTDinterwordspacing

\bibitem{mpc:overviewspringer}
M.~Schwenzer, M.~Ay, T.~Bergs, and D.~Abel, ``Review on model predictive
  control: an engineering perspective,'' \emph{The International Journal of
  Advanced Manufacturing Technology}, vol. 117, no.~5, pp. 1327--1349, 2021.

\bibitem{9812160}
M.~Eckhoff, R.~J. Kirschner, E.~Kern, S.~Abdolshah, and S.~Haddadin, ``An mpc
  framework for planning safe and trustworthy robot motions,'' in \emph{2022
  International Conference on Robotics and Automation (ICRA)}, 2022, pp.
  4737--4742.

\bibitem{britogompc}
B.~Brito, M.~Everett, J.~How, and J.~Alonso-Mora, ``Where to go next: Learning
  a subgoal recommendation policy for navigation among pedestrians,''
  \emph{IEEE Robotics and Automation Letters}, 2021.

\bibitem{td3mpc}
\BIBentryALTinterwordspacing
N.~Hansen, X.~Wang, and H.~Su, ``Temporal difference learning for model
  predictive control,'' 2022. [Online]. Available:
  \url{https://arxiv.org/abs/2203.04955}
\BIBentrySTDinterwordspacing

\bibitem{NEGENBORN2005354}
\BIBentryALTinterwordspacing
R.~R. Negenborn, B.~{De Schutter}, M.~A. Wiering, and H.~Hellendoorn,
  ``Learning-based model predictive control for markov decision processes,''
  \emph{IFAC Proceedings Volumes}, vol.~38, no.~1, pp. 354--359, 2005, 16th
  IFAC World Congress. [Online]. Available:
  \url{https://www.sciencedirect.com/science/article/pii/S1474667016362929}
\BIBentrySTDinterwordspacing

\bibitem{mpc:qlearn}
\BIBentryALTinterwordspacing
M.~Bhardwaj, A.~Handa, D.~Fox, and B.~Boots, ``Information theoretic model
  predictive q-learning,'' 2020. [Online]. Available:
  \url{https://arxiv.org/abs/2001.02153}
\BIBentrySTDinterwordspacing

\bibitem{mpc-greatwood}
C.~Greatwood and A.~G. Richards, ``Reinforcement learning and model predictive
  control for robust embedded quadrotor guidance and control,''
  \emph{Autonomous Robots}, vol.~43, no.~7, pp. 1681--1693, 2019.

\bibitem{ddpgmpc:xue}
\BIBentryALTinterwordspacing
J.~Xue, X.~Kong, B.~Dong, and M.~Xu, ``Multi-agent path planning based on mpc
  and ddpg,'' 2021. [Online]. Available: \url{https://arxiv.org/abs/2102.13283}
\BIBentrySTDinterwordspacing

\bibitem{rlmpc:google}
\BIBentryALTinterwordspacing
T.-Y. Yang, T.~Zhang, L.~Luu, S.~Ha, J.~Tan, and W.~Yu, ``Safe reinforcement
  learning for legged locomotion,'' 2022. [Online]. Available:
  \url{https://arxiv.org/abs/2203.02638}
\BIBentrySTDinterwordspacing

\bibitem{Lange2014ApplicationsOL}
M.~Lange, D.~Z{\"u}hlke, O.~Holz, and T.~Villmann, ``Applications of lp-norms
  and their smooth approximations for gradient based learning vector
  quantization,'' in \emph{ESANN}, 2014.

\bibitem{plappert2018multi}
M.~Plappert, M.~Andrychowicz, A.~Ray, B.~McGrew, B.~Baker, G.~Powell,
  J.~Schneider, J.~Tobin, M.~Chociej, P.~Welinder \emph{et~al.}, ``Multi-goal
  reinforcement learning: Challenging robotics environments and request for
  research,'' \emph{arXiv:1802.09464}, 2018.

\bibitem{todorov2012mujoco}
E.~Todorov, T.~Erez, and Y.~Tassa, ``Mujoco: A physics engine for model-based
  control,'' in \emph{2012 IEEE/RSJ International Conference on Intelligent
  Robots and Systems}.\hskip 1em plus 0.5em minus 0.4em\relax IEEE, 2012, pp.
  5026--5033.

\bibitem{FORCESPro}
\BIBentryALTinterwordspacing
A.~Domahidi and J.~Jerez, ``Forces professional,'' Embotech AG website, 2014.
  [Online]. Available: \url{https://embotech.com/FORCES-Pro}
\BIBentrySTDinterwordspacing

\end{thebibliography}

\balance

\end{document}